\documentclass[preprint,12pt]{elsarticle}

\usepackage{graphicx}
\usepackage{amssymb}

\usepackage{amsmath}
\usepackage{float}
\usepackage{lineno}
\usepackage{subfigure}
\usepackage{hyperref}
\usepackage{color}
\usepackage{multirow}
\modulolinenumbers[5]

\journal{ISPRS}

\begin{document}

\begin{frontmatter}


\title{FCCDN: Feature Constraint Network for VHR Image Change Detection}



\author[1,5]{Pan Chen}
\author[2,3]{Danfeng Hong}
\author[4]{Zhengchao Chen}
\author[1,5]{Xuan Yang}
\author[4]{Baipeng Li}
\author[1,5]{Bing Zhang\corref{mycorrespondingauthor}}
\cortext[mycorrespondingauthor]{Corresponding author at: Aerospace Information Research Institute, Chinese Academy of Sciences, Beijing 100094, China}
\ead{zb@radi.ac.cn}

\address[1]{Key Laboratory of Digital Earth Science, Aerospace Information Research Institute, Chinese Academy of Sciences, Beijing 100094, China}
\address[2]{Remote Sensing Technology Institute (IMF), German Aerospace Center (DLR), 82234 Wessling, Germany}
\address[3]{Data Science in Earth Observation (SiPEO), Technical University of Munich (TUM), 80333 Munich, Germany}
\address[4]{Airborne Remote Sensing Center, Aerospace Information Research Institute, Chinese Academy of Sciences, Beijing 100094, China}
\address[5]{University of Chinese Academy of Sciences, Beijing 100049, China}

\begin{abstract}
\textcolor{blue}{This is a preprint version of a paper submitted to ISPRS Journal of Photogrammetry and Remote Sensing.} 

Change detection is the process of identifying pixelwise differences in bitemporal co-registered images. It is of great significance to Earth observations. Recently, with the emergence of deep learning (DL), the power and feasibility of deep convolutional neural network (CNN)-based methods have been shown in the field of change detection. However, there is still a lack of effective supervision for change feature learning. In this work, a feature constraint change detection network (FCCDN) is proposed. We constrain features both in bitemporal feature extraction and feature fusion. More specifically, we propose a dual encoder-decoder network backbone for the change detection task. At the center of the backbone, we design a nonlocal feature pyramid network to extract and fuse multiscale features. To fuse bitemporal features in a robust way, we build a dense connection-based feature fusion module. Moreover, a self-supervised learning-based strategy is proposed to constrain feature learning. Based on FCCDN, we achieve state-of-the-art performance on two building change detection datasets (LEVIR-CD and WHU). On the LEVIR-CD dataset, we achieve an IoU of an 0.8569 and an F1 score of 0.9229. On the WHU dataset, we achieve an IoU of 0.8820 and an F1 score of 0.9373. Moreover, for the first time, the acquisition of accurate bitemporal semantic segmentation results is achieved without using semantic segmentation labels. This is vital for the application of change detection because it saves the cost of labeling.
\end{abstract}


\end{frontmatter}


\section{Introduction}
\label{S:1}

Change detection is the process of identifying differences in the state of an object or phenomenon by observing it at different times \cite{mahmoudzadeh2007digital}. Since the dynamic monitoring of ground objects is crucial for remote sensing applications, change detection has been the focus and a challenge of remote sensing for a long time. The overall workflow of change detection consists of data acquisition, data preprocessing, a change detection algorithm, and an accuracy evaluation. Traditional change detection algorithms can be divided into algebra-based methods, transformation-based methods, classification-based methods, and clustering-based methods \cite{shi2020change}. Algebra-based methods, including image differing methods \cite{quarmby1989monitoring}, image rationing methods\cite{howarth1981procedures}, and change vector analysis (CVA) \cite{liu2015sequential}, often extract changing information by algebraic operations on the corresponding pixels of bitemporal data. Transformation-based methods detect changes by transforming coregistered images into the feature space. The commonly used transforms include principal component analysis (PCA) \cite{richards1984thematic} and tasseled cap transformation (KT) \cite{jin2005comparison}. Classification-based change detection algorithms obtain a changing area by using the classification results and include postclassification comparisons \cite{ghosh2011fuzzy} and the direct classification of bitemporal data\cite{im2005change}. Clustering-based algorithms generate change maps by clustering bitemporal data into a changed area and an unchanged area. Commonly used clustering algorithms include K-means \cite{liu2019convolutional} and fuzzy c-means (FCM) \cite{cui2019unsupervised}. Although most of the traditional methods are simple and very computationally efficient, their robustness is poor, and their accuracy is not guaranteed.

Deep learning (DL) allows computational models that are composed of multiple processing layers to learn representations of data with multiple levels of abstraction \cite{lecun2015deep}. In recent years, with the development of computing power, the accumulation of data, and proposals of algorithms for mining big data, DL has achieved breakthroughs in many fields. Additionally, remote sensing has entered the era of big data \cite{zhang2019remotely} and has met the high data requirements of DL. Since massive remote sensing data are available, DL can be used to extract useful features and make correct decisions through a large number of remote sensing images, which allows DL-based methods to outperform traditional methods in many remote sensing applications. For the change detection task, a large number of DL change detection algorithms have also been proposed \cite{daudt2018fully,peng2019end,zhang2020deeply,zhang2021object,shi2020change,zheng2021clnet,hou2021high,zhang2021hdfnet}. Early change detection networks are classification networks that input small image patches and output the corresponding categories \cite{zagoruyko2015learning,gao2019change}. With the emergence of fully convolutional networks \cite{long2015fully}, fully convolutional change detection networks have become the preferred architecture \cite{alcantarilla2018street,daudt2018fully,zhang2020deeply,zheng2021clnet,hou2021high,zhang2021hdfnet,fang2021snunet}. Compared with traditional change detection methods, a DL-based algorithm has more hyperparameters, stronger robustness to the input data, and a better generalization ability.

DL architectures make decisions based on the features they learn. Hence, feature learning plays a decisive role in network performance. Currently, Siamese networks with dual encoders and single decoders have been the preferred architecture for very high resolution (VHR) image change detection. They can constrain feature learning by sharing the weights of the dual encoder. Although many Siamese networks have been proposed \cite{alcantarilla2018street,daudt2018fully,zhang2020deeply,diakogiannis2020looking,zhang2021hdfnet,fang2021snunet}, there is still a lack of work on constrained feature learning. The shortcomings can be summarized as follows: 1) Most existing networks focus on the process of extracting and fusing bitemporal features. They tend to ignore some implicit constraints of the change detection task, such as the relationship between bitemporal features. 2) Most existing methods generate change features by fusing encoder features [\cite{daudt2018fully,hou2019w,fang2021snunet}. These low-level features may cause noise inference \cite{zhang2021object}. 3) There is still a lack of an effective solution for the feature misalignment among bitemporal features, especially when the unchanged objects in bitemporal images are very different in the feature space. Therefore, it is still challenging for DL algorithms to extract and fuse bitemporal features correctly.

\begin{figure}[ht]
\centering
\includegraphics[width=0.8\linewidth]{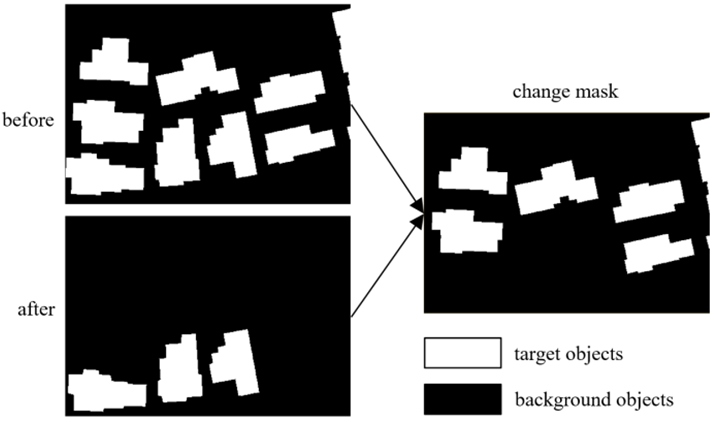}
\caption{Illustration of the change detection task. The change mask can be obtained by comparing bitemporal features.}
\label{fig:change_detection_task}
\end{figure}

Humans can identify changing areas on the basis of understanding the categories of objects in bitemporal images. We can first extract features and classify them into different categories. The category of objects in bitemporal images can help us identify changes. For change detection tasks, extracting the target object features correctly and maintaining a good separation between the target object features and the background features can help the networks perform better. Ideally, we can obtain change information by directly comparing these features (as shown in Figure \ref{fig:change_detection_task}). However, the task of change detection for remote sensing images is extremely complex. On the one hand, the objects in remote sensing images are complex and diverse. It is not easy to distinguish the target objects from the background objects. On the other hand, there are many unconcerned differences among bitemporal data, such as seasonal changes in vegetation and building offsets caused by different viewing angles. These unconcerned differences may result in pseudochanges in the final results.

In this work, we propose a change detection algorithm based on feature constraints. Our work's critical point is to extract the target object features correctly, suppress the background object features, and fuse the bitemporal features more reasonably. We carry out our work from four aspects: a network backbone, multiscale feature extraction and fusion, bitemporal feature fusion, and feature constraints based on self-supervised learning (SSL). Our algorithm is very intuitive and effective, and the performance is very satisfactory.

In summary, the contributions of this work are as follows:

\begin{enumerate}
\item We propose a dual encoder-decoder (DED) backbone for change detection (Section \ref{S:3.2}). Unlike existing works, features from the decoder instead of the encoder are used to calculate change information. In this way, considerable background noise is suppressed and better change features are generated..
\item We design a nonlocal feature pyramid network (NL-FPN) in the center of the backbone to enhance the extraction and fusion of multiscale features (Section \ref{S:3.3}).
\item We design a dense connection-based feature fusion module (DFM) to fuse bitemporal features (Section \ref{S:3.4}). We use two densely connected branches to fuse bitemporal features by ensembling multistage features.
\item We introduce a novel but straightforward SSL task to constrain feature extraction (Section \ref{S:3.5}). The SSL task works by using additional loss on the changed area and unchanged area.
\item We obtain the semantic segmentation results of bitemporal images without semantic segmentation labels on most change detection tasks. The acquisition of the semantic segmentation results is vital for remote sensing because it saves much work with respect to labeling.
\end{enumerate}

We validate our work on two building change detection datasets: LEVIR-CD \cite{chen2020spatial} and WHU \cite{ji2018fully}. The experiments show that our algorithm has obvious advantages over the latest methods. Moreover, the acquisition of the bitemporal building extraction results highlights the significance of our work.

The rest of this paper is organized as follows. Section \ref{S:2} reviews the related work. Section \ref{S:3} introduces our algorithm in detail. Section \ref{S:4} shows our experiments and performance. Section \ref{S:5} discusses the application of the proposed method to multiclass change detection tasks and compares SSL to contrastive loss. Finally, we conclude our work in Section \ref{S:6}.

\section{Related work}
\label{S:2}

\subsection{Fully convolutional networks for change detection}
\label{S:2.1}

FCN is an end-to-end architecture and can make full use of semantic information to obtain pixelwise results. It is naturally suitable for change detection on high-resolution optical images. FCN was first used in street view change detection tasks \cite{alcantarilla2018street}. Rodrigo et al. \cite{daudt2018fully} were some of the first researchers to fit this structure into remote sensing image change detection. They presented three FCNs for change detection tasks and achieved state-of-the-art results on several datasets. 

In recent years, many change detection networks based on FCN have been proposed. FCN-based change detection architectures can be roughly divided into single-stream networks \cite{peng2019end,alcantarilla2018street,peng2020optical,zheng2021clnet,liu2020deep} and double-stream networks \cite{zhang2020deeply,hou2019w,diakogiannis2020looking,zhang2021object,fang2021snunet,zhang2021hdfnet}.

Single-stream networks are usually semantic segmentation networks that take concatenated or differential images of two bitemporal images as input. Peng et al. \cite{peng2019end} use a single-path network to obtain change mask. They first concatenated the coregistered image pairs at the channel dimension. Then, the new multispectral data were fed into a modified UNet++ \cite{zhou2018unet++}. Liu et al. \cite{liu2020deep} inputted concatenated images to a modified UNet \cite{ronneberger2015u}, which was built with depthwise separable convolution. Zheng et al. \cite{zheng2021clnet} proposed a cross-layer convolutional neural network for change detection tasks under the structure of UNet. They designed a cross-layer block to aggregate the multiscale features and multilevel context information.

Double-stream networks are commonly made of two weight-sharing feature extraction streams that directly take bitemporal images as the input. Compared with single-stream backbones, most recent works prefer a double-stream network, owing to the Siamese structure and weight-sharing strategy. Zhang et al. \cite{zhang2020deeply} proposed a fully convolutional two-stream network in which bitemporal features are extracted in a Siamese manner. Then, the paired features are sent to a feature fusion subnetwork to reconstruct the change map. Zhang et al. \cite{zhang2021hdfnet} proposed a Siamese network for change detection tasks with a hierarchical fusion strategy. Bitemporal features are hierarchically fused with concatenating options. Fang et al. \cite{fang2021snunet} extracted bitemporal features with a dual encoder. The bitemporal features are fed into UNet++ to generate change masks. Although bitemporal features are generated by a weight-sharing encoder, features of unchanged objects may vary greatly in bitemporal feature maps. Therefore, how to fuse these features effectively remains a problem. Some researchers fused bitemporal features by concatenation or difference \cite{daudt2018fully,hou2019w,zhang2021hdfnet,fang2021snunet}, while others sought better methods. Chen et al. \cite{chen2020dasnet} proposed a dual attention mechanism that can capture long-range dependencies in feature pairs. The dual attention mechanism is applied before feature fusion to obtain more discriminant feature representations. However, the attention module is only applied to the last stage features of the weight-shared backbone. Since the features are downsampled eight times at the last stage, their method may lose much spatial information. Zhang et al. \cite{zhang2020deeply} introduced an attention module to effectively fuse features in different domains. The attention module they built is a combination of a channel attention module and a spatial attention module. Diakogiannis et al. \cite{diakogiannis2020looking} designed a new attention module to calculate change features. The module uses the fractal Tanimoto similarity to compare queries with keys inside the attention module. Zhang et al. \cite{zhang2021object} proposed an object-level change detection network with a dual correlation attention-guided detector. They built a correlation attention-guided feature fusion neck that uses a position-correlated attention module and channel-correlated attention module to guide the generation of change features. These methods use a series of attention modules that fuse bitemporal features hierarchically. Although they can remarkably boost the performance of networks, most attention-based fusion methods introduce a large amount of computation and a large number of parameters. Moreover, there still exist several issues that need to be considered. On the one hand, most existing works fuse shallow features to obtain the change information, which may introduce considerable noise in change maps \cite{zhang2021object}. On the other hand, existing change detection networks mainly rely on the supervision of labels. They tend to ignore the relationship among bitemporal features. Inspired by the above issues, a feature constraint architecture for change detection tasks is proposed in this literature.

\subsection{Self-supervised learning}
\label{S:2.2}

SSL is a subset unsupervised learning method that can learn general image features without using any human-annotated labels \cite{jing2020self}. It is widely used in the computer vision field. Unlike other unsupervised learning methods, SSL trains with annotations, which are obtained by mining the internal information of images through certain pretext tasks. There have been a variety of pretext tasks for SSL, such as colorizing grayscale images \cite{zhang2016colorful}, image inpainting \cite{pathak2016context}, image rotation \cite{feng2019self}, and image jigsaw puzzles \cite{noroozi2016unsupervised}.

Recently, several researchers in the field of remote sensing have carried out their research with SSL. Tao et al. \cite{tao2020remote} introduced a new instance discrimination-based SSL mechanism to pretrain a feature exactor for remote sensing scene classification tasks. They also investigated the impacts of several factors in the SSL-based classification task. Vincenzi et al. \cite{vincenzi2020color} took advantage of SSL and trained a promising feature extractor for the downstream landcover classification task. Their pretext task was to reconstruct the visible colors with high-dimensionality spectral bands. For change detection, Dong et al. \cite{dong2020self} an SSL-based change detection algorithm. They built a network to identify different sample patches between two temporal images, namely, temporal prediction. With this pretext task, the network can encode input images to more consistent feature representations, which can be used to generate binary change maps by clustering. Chen et al. \cite{chen2021self} proposed an SSL-based change detection approach based on an unlabeled multiview setting. They used SSL by pretraining a Siamese network with a contrastive loss for heterogeneous images and a regression loss for homogeneous images. Although these change detection tasks benefit from the SSL-based strategy, SSL is only used as a pretraining strategy. There is still a lack of research on the introduction of an SSL strategy to change detection tasks.

\section{Methodology}
\label{S:3}

In this section, we introduce the architecture of the proposed method in detail. The overall structure of FCCDN is presented in Section \ref{S:3.1}. More specifically, we propose a novel DED backbone for the change detection task (Section \ref{S:3.2}), an NL-FPN module for multiscale feature extraction (Section \ref{S:3.3}), a DFM block for generating change features (Section \ref{S:3.4}), and a feature constraint strategy based on SSL (Section \ref{S:3.5}). 

\subsection{Overall structure of proposed network architecture}
\label{S:3.1}
\begin{figure}[ht]
\centering
\includegraphics[width=1\linewidth]{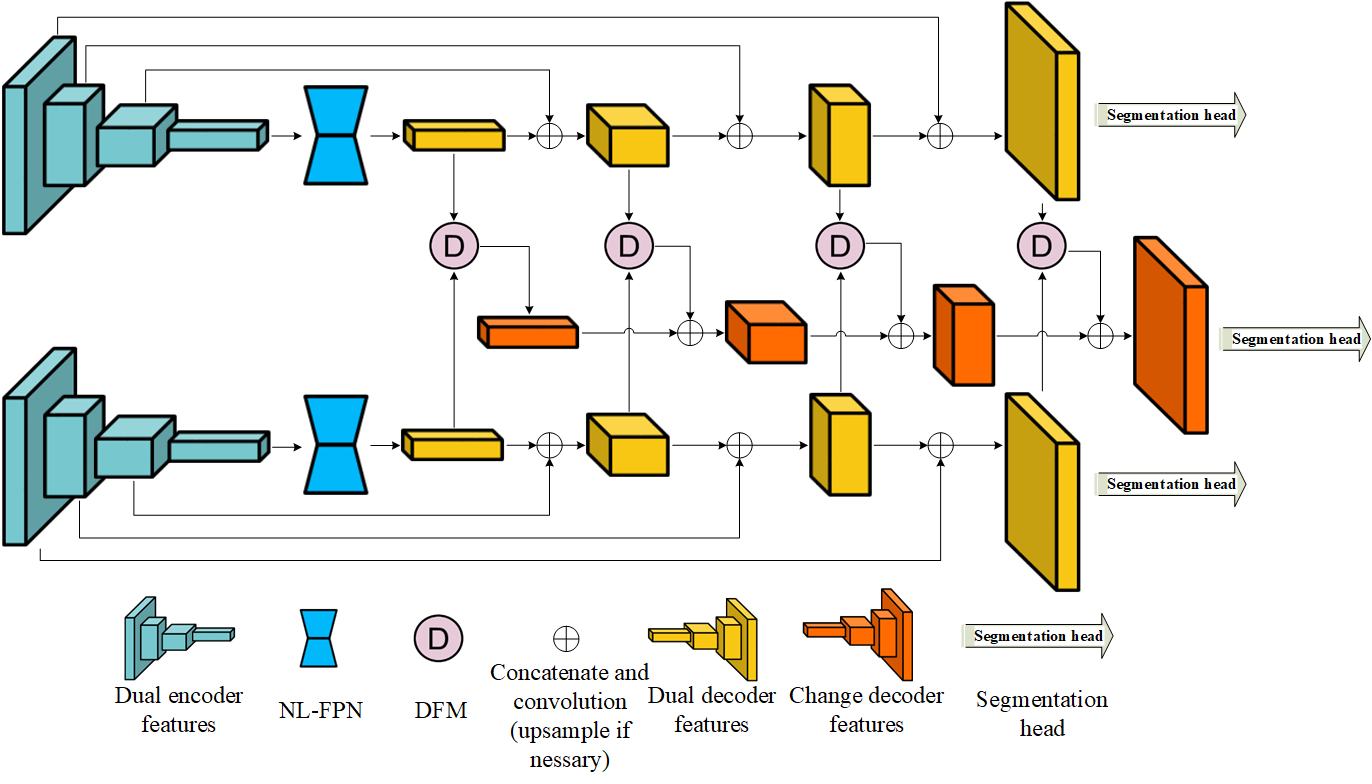}
\caption{The overall structure of FCCDN.}
\label{fig:overall}
\end{figure}

The overall structure of the proposed feature constraint change detection network (FCCDN) is shown in Figure \ref{fig:overall}. FCCDN uses a DED backbone (Figure \ref{fig:DED}). At the center of DED, an NL-FPN (Figure \ref{fig:NL_FPN}) is proposed to enhance multiscale features in a nonlocal way. At the decoding stage, a series of DFMs (Figure \ref{fig:DFM}) are applied to fuse bitemporal features and generate change features. Then, the change features are hierarchically fed to a change decoder to obtain the final change map. FCCDN contains three output branches: one change branch and two segmentation branches Figure \ref{fig:SSL}). The change branch produces the change confidence, which is used to calculate the change loss with labels in the training process. The two segmentation branches output segmentation scores and are supervised by the SSL-based strategy.

\subsection{Dual encoder-decoder network}
\label{S:3.2}

Since bitemporal data are usually used in change detection tasks, a fully convolutional Siamese (FCS) network has been a common structure for change detection in recent years. Conventional FCS (also referred to as FCS) generally uses a dual encoder with weight sharing to extract bitemporal features. After the encoding process, a series of feature fusion blocks are applied to obtain change features based on the extracted features. At the decoding stage, a single-stream change decoder is used to combine the change features mentioned above in a similar way as UNet \cite{ronneberger2015u}.

\begin{figure}[H]
\vspace{-0.35cm}
\centering
\subfigure[]{
    \includegraphics[width=0.5\linewidth]{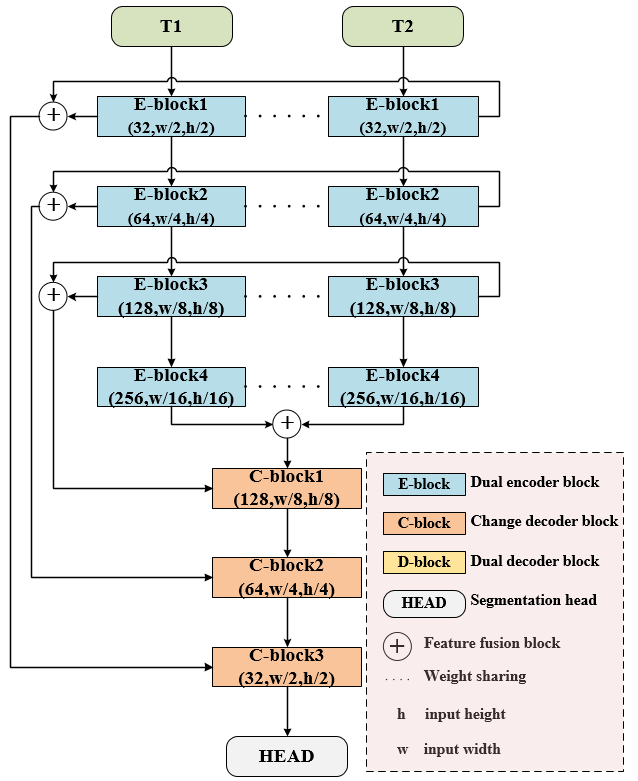}
    \label{fig:FCS}
}
\hfill
\vspace{-0.35cm}
\subfigure[]{
    \includegraphics[width=0.4\linewidth]{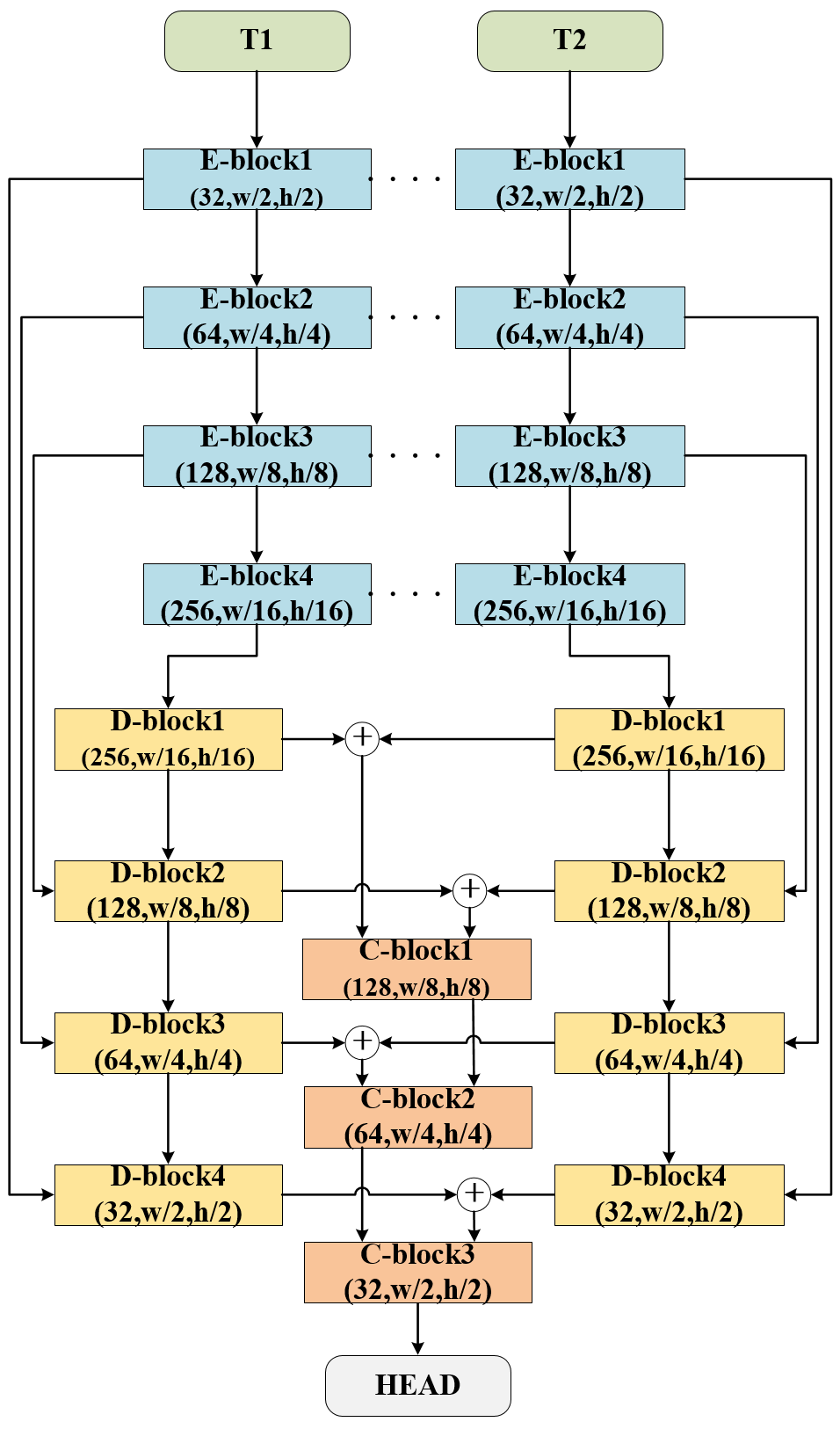}
    \label{fig:DED}
}
\hfill
\vfill
\subfigure[]{
    \includegraphics[width=0.5\linewidth]{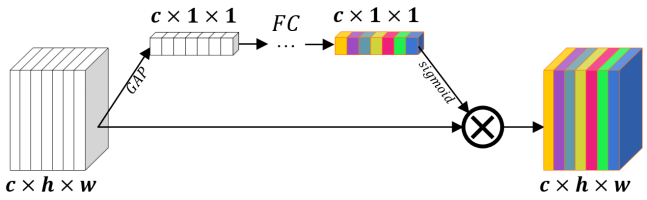}
    \label{fig:SE}
}
\vfill
\vspace{-0.35cm}
\subfigure[]{
    \includegraphics[width=0.35\linewidth]{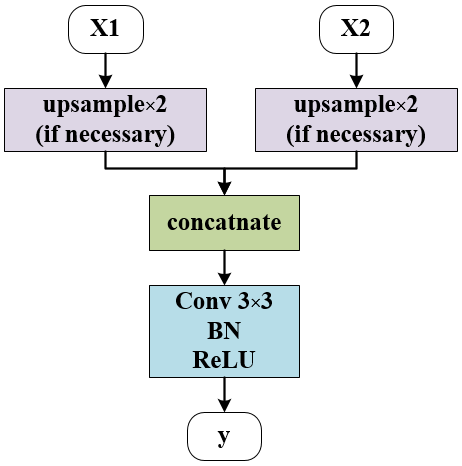}
    \label{fig:cat}
}
\hfill
\vspace{-0.35cm}
\subfigure[]{
    \includegraphics[width=0.35\linewidth]{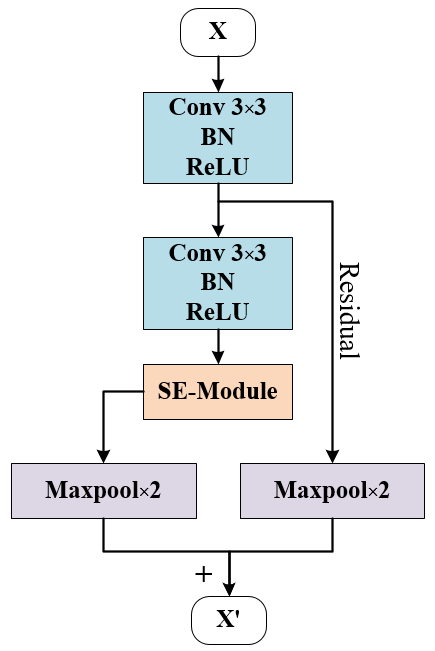}
    \label{fig:SE-Res}
}
\caption{The backbone architecture for the change detection task. (a) The details of FCS.  (b) The details of DED. (c) The SE block in the SE-ResNet module. (d) The module we used to build bitemporal feature fusion module blocks and decoder blocks. (e) The SE-ResNet module we use to build the encoder.}
\label{fig:backbone}
\end{figure}

Figure \ref{fig:FCS} shows the overall structure of our baseline FCS. We build the dual encoder with a series of SE-ResNet modules (Figure \ref{fig:SE-Res}) \cite{hu2018squeeze}, which uses the squeeze and excitation (SE) block (Figure \ref{fig:SE}) in the residual branch. The SE block uses global average pooling (GAP) to aggregate the feature maps across spatial dimensions and captures channelwise feature responses. After GAP, two fully connected (FC) layers are applied to learn a weight vector for each channel. The vector is then normalized and multiplied by the original feature. The dual encoder is built with four SE-ResNet modules. The output of the four modules are features with sizes of (h/2, w/2), (h/4, w/4), (h/8, w/8), and (h/16, w/16) (h represents the height of the input image and w represents the width of the input image). The encoder features are then fed to four feature fusion modules to fuse bitemporal features and generate four change features. In the decoding process, several change decoder blocks are employed to fuse features generated by bitemporal feature fusion blocks hierarchically. Since the bitemporal feature fusion block and the change decoder block are both aimed at fusing features, we build them with the same module, as shown in Figure \ref{fig:cat}. This fusion module is a combination of an upsampling layer (if necessary), a concatenate operation, a convolution layer with kernel size = 3, a batch normalization (BN) layer \cite{ioffe2015batch}, and a rectified linear unit (ReLU) function. The output of the last change decoder block (C-block3) is then fed into a simple segmentation head, which is meant to reduce the channels and resize the change mask to the input size.

\begin{figure}[!ht]
\vspace{-0.35cm}
\centering
\subfigure[]{
    \includegraphics[width=0.45\linewidth]{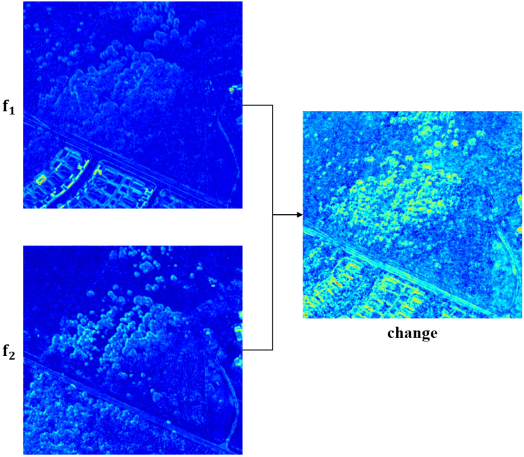}
    \label{fig:fm_FCS}
}
\hfill
\vspace{-0.35cm}
\subfigure[]{
    \includegraphics[width=0.45\linewidth]{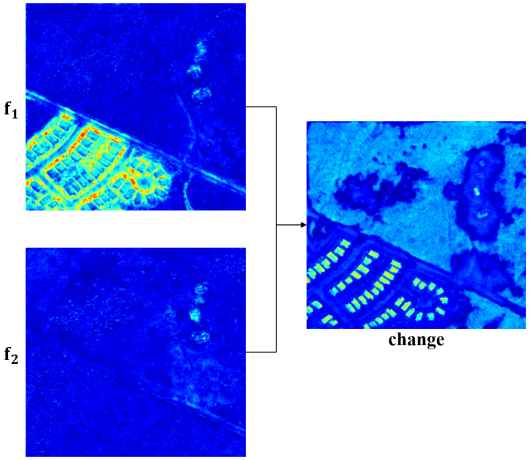}
    \label{fig:fm_DED}
}
\caption{Comparison of features in FCS and DED. $f_{\rm 1}$ and $f_{\rm 2}$ refer to bitemporal features. change refers to change features. (a) Bitemporal features and change features from the dual encoder of FCS. (b) Bitemporal features and change features from the dual decoder of DED.}
\label{fig:fm_FCS_DED}
\end{figure}

The conventional FCS network uses low-level features in the dual encoder to calculate the change features. However, low-level features pay more attention to general information, which contains much information that is useless for change detection \cite{zhang2021object}. We validate this conjecture with the change detection task on buildings and visualize some features in Figure \ref{fig:fm_FCS_DED}. The feature maps are visualized in the form of heatmaps with Grad-CAM \cite{selvaraju2017grad}. In the heatmap, the warmer the color is, the more attention the network pays to that area. That is, the degree of warmth represents the contribution of each pixel. As shown in Figure \ref{fig:fm_FCS}, the low-level features ($f_{\rm 1}$ and $f_{\rm 2}$) from the dual encoder contain not only the edges of buildings but also the edges of background objects. In this case, there is much noise in the change features, which is not conducive to the change detection task.

In this paper, we propose a novel DED neural network backbone for change detection tasks (see Figure \ref{fig:DED}). This architecture has the same dual encoder and change decoder as FCS. Nevertheless, unlike FCS, DED has an extra dual decoder with weight sharing, which is used to filter out invalid information and reconstruct useful information. The dual decoder is also built with the feature fusion block shown in Figure \ref{fig:cat}. Change features are generated by fusing the corresponding features from the dual decoder. Then, these change features are fed into a change decoder in the same way as in FCS. The bitemporal features reconstructed by the dual decoder are shown in Figure \ref{fig:fm_DED}. Compared with the bitemporal features from the dual encoder of FCS, the bitemporal features from the dual decoder can better highlight the borders of buildings and suppress background features. Therefore, the change features generated by DED are more accurate.

\subsection{Non-local feature pyramid network}
\label{S:3.3}

It is well known that ground objects vary on different scales. Therefore, it is necessary to extract and fuse features at multiple scales and with large receptive fields. Some researchers try to tackle the above issue with pyramid pooling (PSP) \cite{zhao2017pyramid,diakogiannis2020resunet}] or atrous spatial pyramid pooling (ASPP) \cite{chen2017rethinking}. However, these methods use large-scale pooling or convolution layers with a large dilate rate, resulting in the absence of local information \cite{li2018pyramid}. In addition, there are also some solutions based on a feature pyramid network (FPN) \cite{lin2017feature,mi2020superpixel}, which can effectively fuse multiscale features. However, conventional FPN methods cannot obtain long-range dependencies, which makes it hard to extract large object features.

\begin{figure}[!ht]
\vspace{-0.35cm}
\centering
\subfigure[]{
    \includegraphics[width=0.45\linewidth]{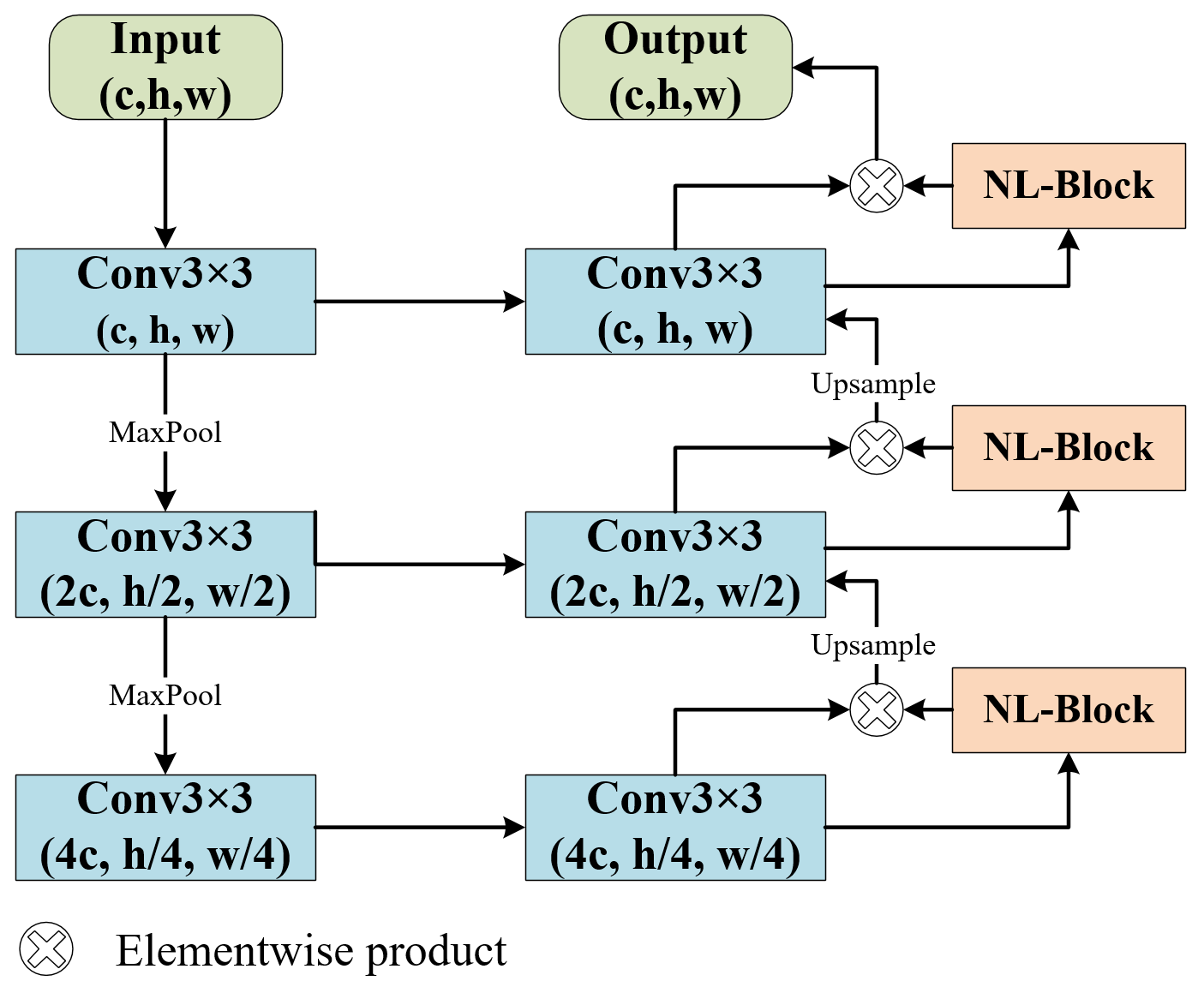}
    \label{fig:FPN}
}
\hfill
\vspace{-0.35cm}
\subfigure[]{
    \includegraphics[width=0.45\linewidth]{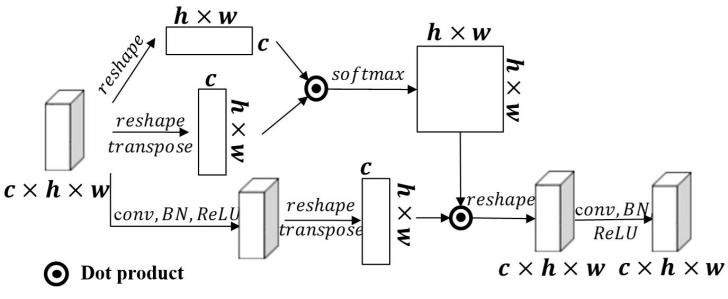}
    \label{fig:NL_block}
}
\caption{The illustration of NL-FPN. (a) The overall structure of the NL-FPN. (b) The details of NL-block.}
\label{fig:NL_FPN}
\end{figure}

Here, we propose an NL-FPN module to extract and fuse features in a nonlocal way. The module is added to the center of FCCDN (Figure \ref{fig:overall}) to augment the feature maps by considering similarities between any pairs of pixels. Because similarities between any two positions is considered, we can further strengthen the intraclass correlations and increase interclass separation. Figure \ref{fig:NL_FPN} shows the details of our NL-FPN. Nonlocal blocks (NL-blocks) are added to the upsampling stage of FPN, which consists of six convolution layers. We build the NL-block inspired by the spatial self-attention mechanism \cite{wang2018non,yuan2018ocnet}. The self-attention mechanism uses the dot product between the key vector and the query vector to obtain the similarity between different pixels. Since the features are normalized to unit lengths, the distance between vectors can be computed using the dot product \cite{xu2017accurate}. We can formulate this as

\begin{equation}
\label{eq:nl_block}
1-\left ( F_{p} \right )^{T}F^{q}=\frac{1}{2}\left \| F^{p}-F^{q} \right \|^{2}
\end{equation}

\noindent
where F represents the feature map and p and q represent different positions on the feature map. For this reason, we directly consider the dot product of the original input and its transpose instead of using a key vector and query vector. In this way, the similarity can also be obtained, and the network parameters can be reduced.

The NL-block we build is shown in Figure \ref{fig:NL_block}. The input feature map is fed into three branches: a reshape branch, a reshape and transpose branch, and a convolution branch. The features in the first two branches are fused by a dot product followed by a softmax function. Here, a similarity map is generated. The feature in the convolution branch is fed into a convolution layer. Then, the output of the convolution branch is reshaped and transposed so that we can further fuse it with the similarity map mentioned before by using a dot product. At the end of this block, the fusion result is reshaped and fed into a convolution branch to obtain the final weight map. The weight map is used to augment the features in FPN by a Hadamard product.

\subsection{Dense fusion module}
\label{S:3.4}

For change detection with Siamese networks, bitemporal feature fusion is the most critical part. It is difficult for two reasons: (1) Bitemporal images fed into Siamese networks are often offset in spatial position and color. (2) The background objects are complex and diverse. Conventional methods use direct subtraction or concatenation to fuse features \cite{daudt2018fully}. Unfortunately, although Siamese networks extract features by dual blocks, there is still much misalignment among the bitemporal features. There have also been many researchers who have tried to tackle this problem with attention mechanisms \cite{zhang2020deeply,chen2020dasnet,diakogiannis2020looking,zhang2021object}. Nevertheless, most existing attention-based feature modules introduce many calculations and consume considerable memory.

\begin{figure}[ht]
\centering
\includegraphics[width=0.8\linewidth]{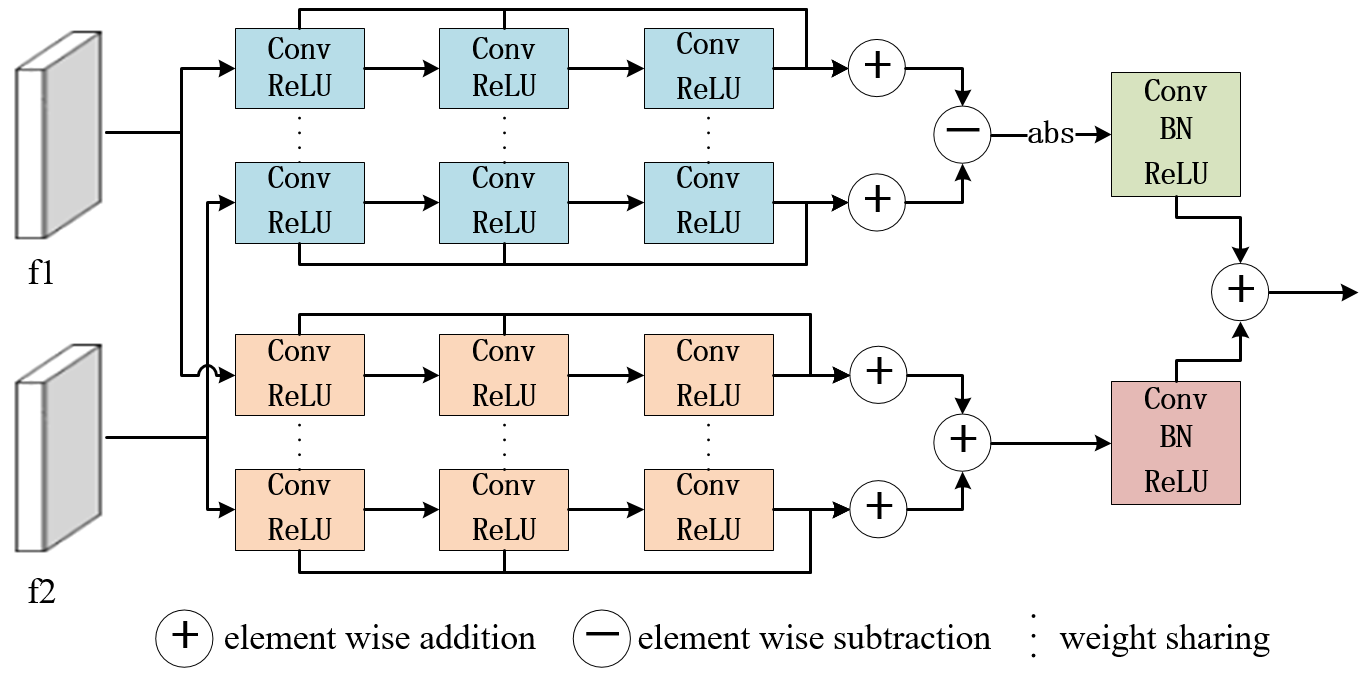}
\caption{Illustration of DFM.}
\label{fig:DFM}
\end{figure}

Here, we propose a simple yet effective feature fusion module based on dense connections. We name this module DFM. DFM consists of two branches, the sum branch and the difference branch. The sum branch is used to enhance the edge information, and the difference branch is used to generate change regions. Each branch is built with two densely connected streams with weight sharing. Figure \ref{fig:DFM} illustrates the details of DFM. All convolution operations use 3×3 kernels. We should note that we do not use BN in each branch until the final convolution layer.

\begin{figure}[ht]
\centering
\includegraphics[width=0.8\linewidth]{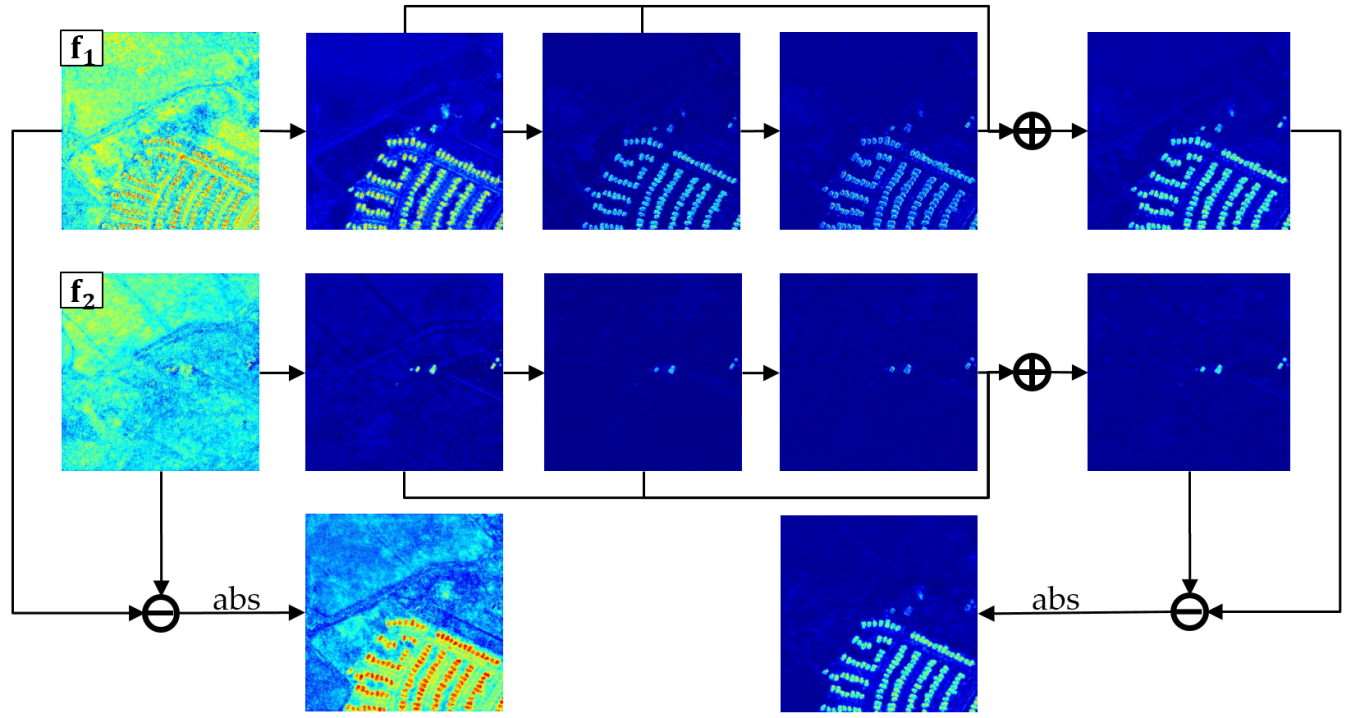}
\caption{Visualization of features in the difference branch. $f_{\rm 1}$ and $f_{\rm 2}$ denote the bitemporal features fed into this branch. The first two rows show the features generated by the dual dense connected stream. On the left of the third row is the change feature generated before the difference branch. On the right of the third row is the change feature generated after the difference branch.}
\label{fig:fm_DFM}
\end{figure}

We build DFM for the purpose of ensembling multiple features in each stream and making better decisions. This structure can increase the robustness of the model and prevent pseudochanges caused by feature misalignment. Additionally, owing to the rich residual connections in dense connections, the last two features in each stream can be regarded as the residual of the previous feature, which is to some extent the correction of the previous feature and makes the new feature map more aligned. We validate this module by visualizing features in the difference branch (Figure \ref{fig:fm_DFM}). From the visualization of the feature maps, we can see that DFM indeed reduces feature misalignment and calculates more accurate change features.

\subsection{Self-supervised learning-based feature constraint}
\label{S:3.5}

Change detection is aimed at finding changes of interest in bitemporal images. Currently, most change detection tasks are inter-class change detection, in which the categories of bitemporal images should differ in changed areas, and the categories in the unchanged areas should be the same. That is, the bitemporal features in unchanged areas should be as close as possible, and the bitemporal features in changed areas should be as far away as possible. To this end, we use the novel idea of applying SSL to the task of change detection. With this SSL-based strategy, we can further constrain feature learning. 

\begin{figure}[ht]
\centering
\includegraphics[width=0.8\linewidth]{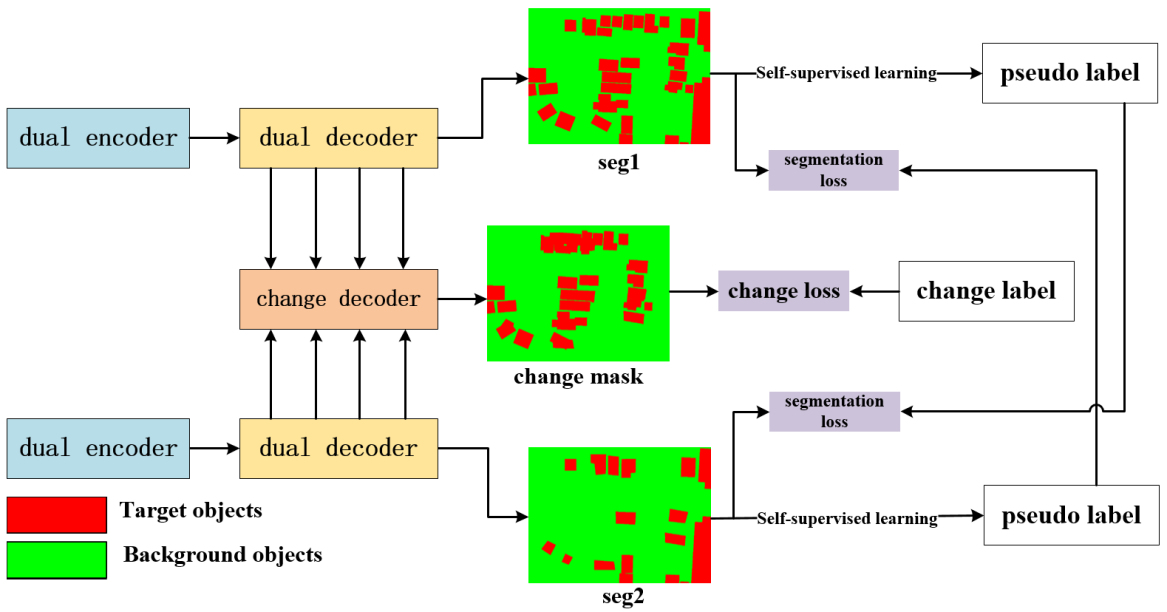}
\caption{The overall workflow of SSL-based feature constrains.}
\label{fig:SSL}
\end{figure}

As shown in Figure \ref{fig:SSL}, we add two auxiliary branches to the dual decoder of the proposed DED backbone. The two auxiliary branches are meant to obtain the semantic segmentation results of target objects. For the learning of the auxiliary branches, we use a novel SSL strategy to generate pseudolabels for each branch. It should be noted that we take the binary change detection task as an example to describe our idea. We discuss the application of the SSL-based feature constraint strategy for multiclass change detection tasks in the discussion section(Section \ref{S:5}). The details of the SSL strategy are as follows: (1) According to the change detection label, we split the semantic segmentation results of the auxiliary branches ($Seg_{\rm 1}$ and $Seg_{\rm 2}$ in Figure \ref{fig:SSL}) into two parts: the changed area and unchanged area. (2) In the unchanged area, the semantic segmentation result of one branch is used as the label of the other branch. (3) In the changed area, the opposite semantic segmentation result of one branch is used as the label of the other branch.

The way we generate pseudolabels can be expressed by the following formulas. We assume that the output of the two auxiliary branches is $S_{\rm 1}$, $S_{\rm 2} \in R ^ {x\times y\times 1}$. Since $S_{\rm 1}$ and $S_{\rm 2}$ are normalized to [0, 1], we can obtain the semantic segmentation results of the target object ($P_{\rm 1}$ and $P_{\rm 2}$) with:

\begin{equation}
\begin{aligned}
P_{1,i} &= \left\{\begin{matrix}
0 & S_{1,i}< 0.5\\ 
1 & S_{1,i}\geqslant 0.5
\end{matrix}\right.
& P_{2,i} &= \left\{\begin{matrix}
0 & S_{2,i}< 0.5\\ 
1 & S_{2,i}\geqslant 0.5
\end{matrix}\right.
\end{aligned}
\label{eq:get_pred}
\end{equation}

\noindent
where i represents the position of each pixel. Due to $P_{\rm 1}$, $P_{\rm 2}\in\left \{0, 1 \right \}$, based on the idea of SSL, we use them as pseudolabels to constrain each semantic segmentation branch. For example, we use $P_{\rm 2}$ as the label of $P_{\rm 1}$. As mentioned earlier, the categories of bitemporal images in the changed area must be different, and the categories in the unchanged area must be the same. We split $P_{\rm \*}$ and $P_{\rm \*}$ into two parts: the changed area C and the unchanged area U. Then, we feed semantic segmentation results and pseudolabels into loss functions. Hence, we obtain: 

\begin{equation}
\begin{aligned}
l_{seg1}=F\left (S_{1,i},P_{2,i} | i\in U\right )+F\left (S_{1,i},1-P_{2,i} | i\in C\right ) \\
l_{seg2}=F\left (S_{2,i},P_{1,i} | i\in U\right )+F\left (S_{2,i},1-P_{1,i} | i\in C\right )
\end{aligned}
\label{eq:ssl_loss}
\end{equation}

\noindent
where F denotes the loss function, $l_{\rm seg1}$ and $l_{\rm seg2}$ denote the auxiliary losses we introduce. In the end, our loss of change detection task consists of three parts: a loss of change mask ($l_{\rm change}$), pretemporal semantic segmentation loss ($l_{\rm seg1}$), bitemporal semantic segmentation loss ($l_{\rm seg2}$). The final loss function can be formulated as:

\begin{equation}
loss=l_{change}+0.2*l_{seg1}+0.2*l_{seg2}
\label{eq:final_loss}
\end{equation}

Both the change loss and auxiliary loss are calculated with the same loss function, which is a combination of the binary cross-entropy (BCE) loss and dice coefficient loss functions \cite{zhou2018d}. The BCE loss can be formulated as follow:

\begin{equation}
\begin{aligned}
l_{BCE}=\frac{1}{N}\sum_{n=1}^{N}\left ( y_{n}\log x_{n} +\left ( 1-y_{n} \right )\log \left ( 1-x_{n} \right )\right )
\end{aligned}
\label{eq:bce_loss}
\end{equation}

where N represents the total number of pixels in a label patch $x_{\rm n}$ and $y_{\rm n}$  denote the predicted change confidence and the label in the corresponding position, respectively. The dice coefficient loss is calculated as:

\begin{equation}
\begin{aligned}
l_{dice}=1-\frac{2*\left | X\bigcap Y \right |}{\left | X \right |+\left | Y \right |}
\end{aligned}
\label{eq:dice_loss}
\end{equation}

where X and Y represent the predicted change confidence and the label $\cap$ denotes the intersection of X and Y. Hence, the BCE + dice loss we use can be expressed as:

\begin{equation}
\begin{aligned}
l=l_{BCE}+l_{dice}
\end{aligned}
\label{eq:BCEdice_loss}
\end{equation}

\begin{figure}[ht]
\centering
\includegraphics[width=0.8\linewidth]{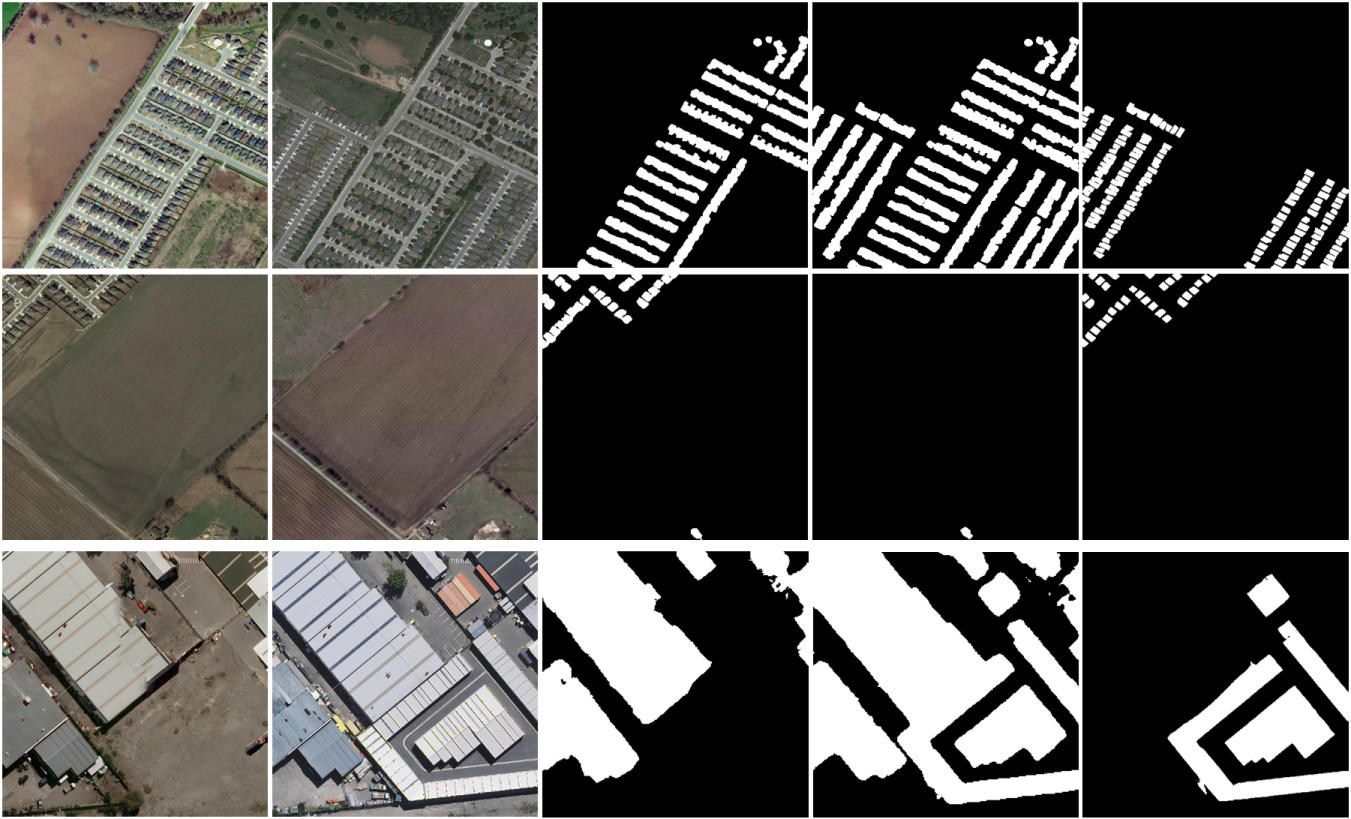}
\caption{Examples of the change detection results and bitemporal semantic segmentation results. (a) pretemporal images. (b) posttemporal images. (c) pretemporal semantic segmentation results. (d) posttemporal semantic segmentation results. (e) change masks.}
\label{fig:show_SSL}
\end{figure}

There is a simple way to validate our SSL-based feature constraint strategy. That is, we can validate the improvement by checking the inference results of the two auxiliary semantic segmentation branches. The two auxiliary branches are meant to guide DED to extract better target object features and suppress background object features. Therefore, the accuracy of the semantic segmentation results directly represents the performance of our SSL-based feature constraint. We show several results in Figure \ref{fig:show_SSL}. From left to right: pretemporal images, posttemporal images, the pretemporal semantic segmentation results, the posttemporal semantic segmentation results, and the change detection results. Apparently, the two semantic segmentation branches achieve good performance and further improve the change task. 

Note that we only test the SSL-based strategy on inter-class change detection tasks since almost all the available change detection datasets are inter-class change detection. It is hard to tell whether this strategy can boost performance on intra-class change detection tasks. So, we can only conclude that the SSL-based strategy can effectively constrain the feature learning of inter-class change detection tasks.

\section{Experimental Results}
\label{S:4}

We validate FCCDN on two building change detection datasets: LEVIR-CD and WHU. The experimental results demonstrate that FCCDN outperforms recently proposed change detection methods and achieves state-of-the-art performance on experimental datasets. In this section, we start by introducing the experimental datasets. Then, we describe our implementation details. After that, we introduce the evaluation metrics we use. In the end, we present our results in detail.

\subsection{Datasets}
\label{S:4.1}

We describe the experimental datasets in this subsection. We offer a brief view of the LEVIR-CD dataset and WHU dataset in Table \ref{Table:datasets}. More details are shown in Section \ref{S:4.1.1} and Section \ref{S:4.1.2}.

\begin{table}[ht]
\centering
\begin{tabular}{l c c c c}
\hline
\textbf{Name} & \textbf{Bands} & \textbf{Image pairs} & \textbf{Resolution(m)} & \textbf{Image size}\\
\hline
WHU & 3 & 1 & 0.3 & 32207×15354 \\
LEVIR-CD & 3 & 637 & 0.5 & 1024 × 1024 \\
\hline
\end{tabular}
\caption{A brief introduction of the WHU dataset and LEVIR-CD dataset.}
\label{Table:datasets}
\end{table}

\subsubsection{LEVIR-CD dataset}
\label{S:4.1.1}

The LEVIR-CD dataset consists of 637 VHR image patches collected from Google Earth (GE). The resolution of each image is 0.5 m, and the size is 1024×1024. It is a large-scale change detection dataset and covers different kinds of buildings. The author of LEVIR-CD provided a standard train/validation/test split, which assigns 70\% of the samples for training, 10\% for validation, and 20\% for testing. We follow the standard split provided by the author. Most existing literature crops the samples into 256 × 256 \cite{chen2020spatial,diakogiannis2020looking,peng2020optical}. Theoretically, large training slices contain more context information than small training slices. Owing to the low computational cost of FCCDN, we can feed larger slices into the network. We try two crop strategies: 1) we crop the samples into 256 × 256 with an overlap of 128 on each side (horizontal and vertical); 2) we crop the samples into 512 × 512 with an overlap of 256 on each side.

\subsubsection{WHU building change detection dataset}
\label{S:4.1.2}

The WHU building change detection dataset consists of two-period aerial images, each with a resolution of 0.3 m. The two-period images were obtained in 2012 and 2016. There are a variety of buildings with large-scale changes in the dataset. There does not exist a standard splitting for this dataset. Different researchers use different data splitting approaches to validate their models. For the convenience of comparison, we use the splitting approach that was used in \cite{peng2020optical}, and several change detection architectures have been tested based on this splitting approach. We crop the dataset into 256 × 256 slices and randomly split them into training/validation/test sets at a ratio of 7:1:2. Note that we do not use any overlap during the splitting.

\subsection{Implementation Details}
\label{S:4.2}

\subsubsection{Data preprocessing and augmentation}
\label{S:4.2.1}

We generate the training set and validation set in the way mentioned in Section \ref{S:4.1}. For each dataset, the slices are normalized according to Equation \ref{eq:input_normalize} before being fed into the network.

\begin{equation}
img^{'}=\frac{img-mean}{std}
\label{eq:input_normalize}
\end{equation}

\noindent
where $img$ represents the slices before normalization, ${img}'$ represents the slices after normalization, and $mean$ and $stdv$ are the mean value and standard deviation of the images in the datasets, respectively.

To improve the generalization ability of the models, we use several data augmentation strategies in the training stage. This includes random flipping (probability = 0.5), transposing (probability = 0.5), rotating (probability = 0.3, -45° $\leq$ angle $\leq$ 45°), zooming in/out (probability = 0.3, scale $\leq$ 0.1), HSV shifting (probability = 0.3, H-shift $\leq$ 10, S-shift $\leq$ 5, V-shift $\leq$ 10), and adding Gaussian noise (probability = 0.3, mean = 0, 10 $\leq$ variance $\leq$ 50). All the above data augmentation methods are realized with Albumentations \cite{buslaev2020albumentations}, which is a Python library for data augmentation. In addition, we randomly exchange the input order of the bitemporal images (probability = 0.5).

\subsubsection{Training}
\label{S:4.2.2}

FCCDN is implemented with the PyTorch DL framework \cite{paszke2017automatic}. We have open-source our work on GitHub, and here is the link: \url{https://github.com/chenpan0615/FCCDN_pytorch}. We trained our networks on 1 RTX TITAN GPU (24 GB memory). For the proposed network architecture, the minibatch size can reach 16 with 512 × 512 slices (LEVIR-CD) and 64 with 256 × 256 slices (WHU). We choose the BCE + dice coefficient loss as the loss function. We use AdamW \cite{kingma2014adam} as the optimizer with an original learning rate = 0.002 and weight decay = 0.001. The learning rate is adjusted by observing whether the F1 score of the validation set increases within ten epochs. If there is no increase, the learning rate is reduced by a factor of 0.3. We find that the models almost converge when the learning rate is adjusted more than three times. Thus, we end the training process when the learning rate is about to adjust a 4th time. Since the models are far from converging in the first 30 epochs, we do not validate the models with the validation sets until the 30th epoch. The pretrained model is vital for improving model robustness and uncertainty \cite{hendrycks2019using}, especially for training with a small training set. Therefore, in the experiment on the WHU dataset, we initialize the encoder of FCCDN with the weights trained on the LEVIR-CD dataset.

\begin{figure}[H]
\centering
\vspace{-0.35cm}
\includegraphics[width=.45\linewidth]{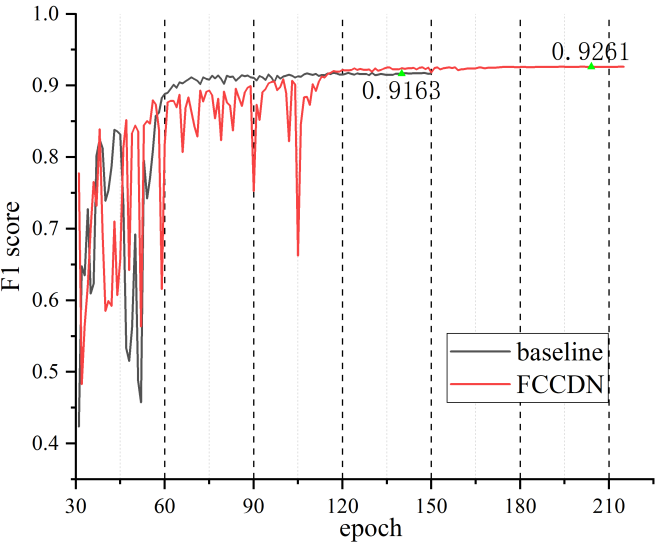}\hfill \hspace{-0.45cm}
\includegraphics[width=.45\linewidth]{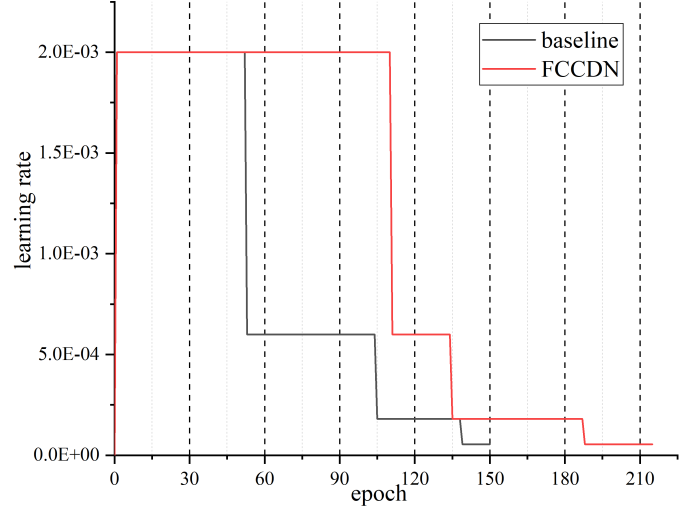}\\[0.5mm]
\includegraphics[width=.45\linewidth]{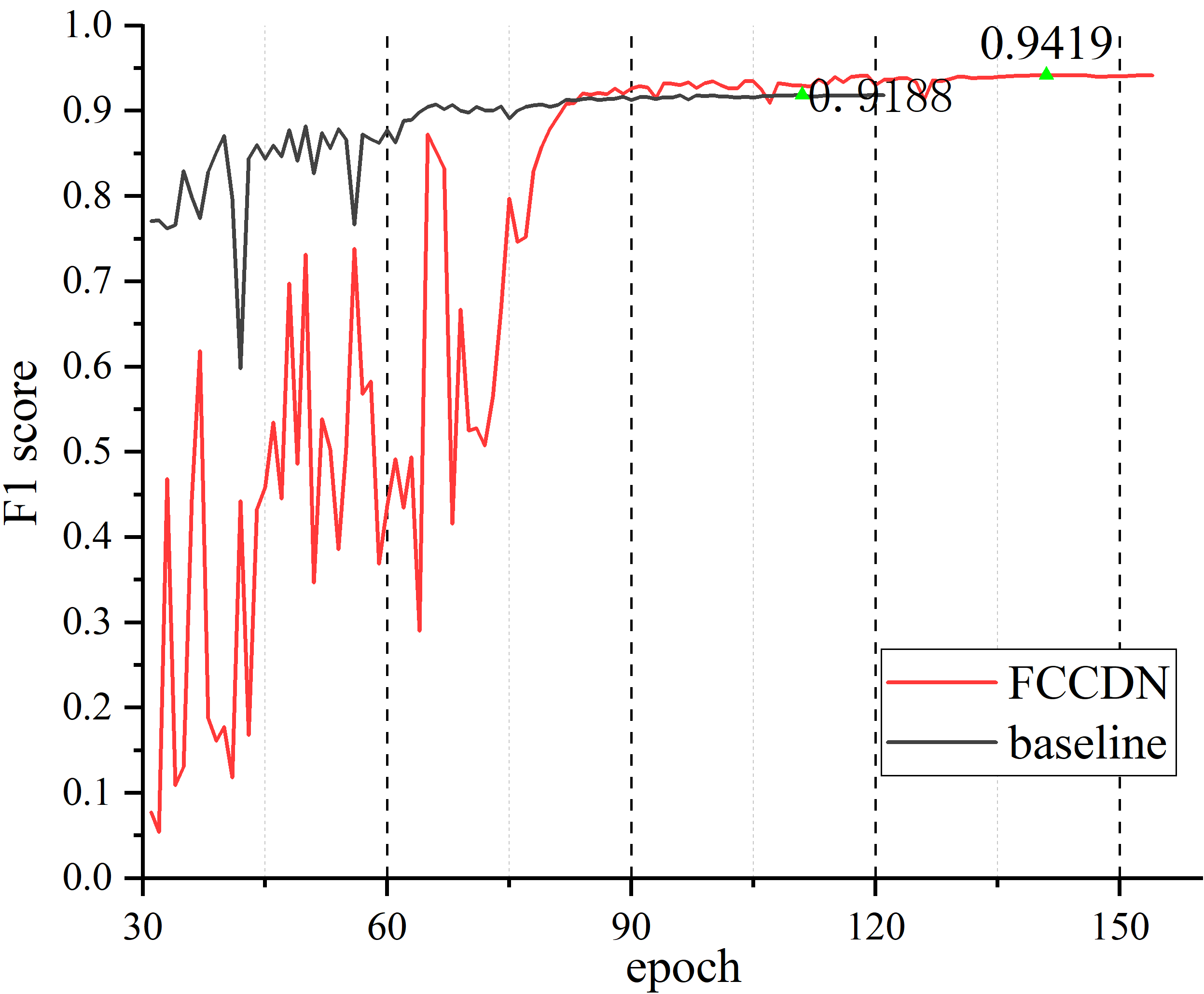}\hfill \hspace{-0.45cm}
\includegraphics[width=.45\linewidth]{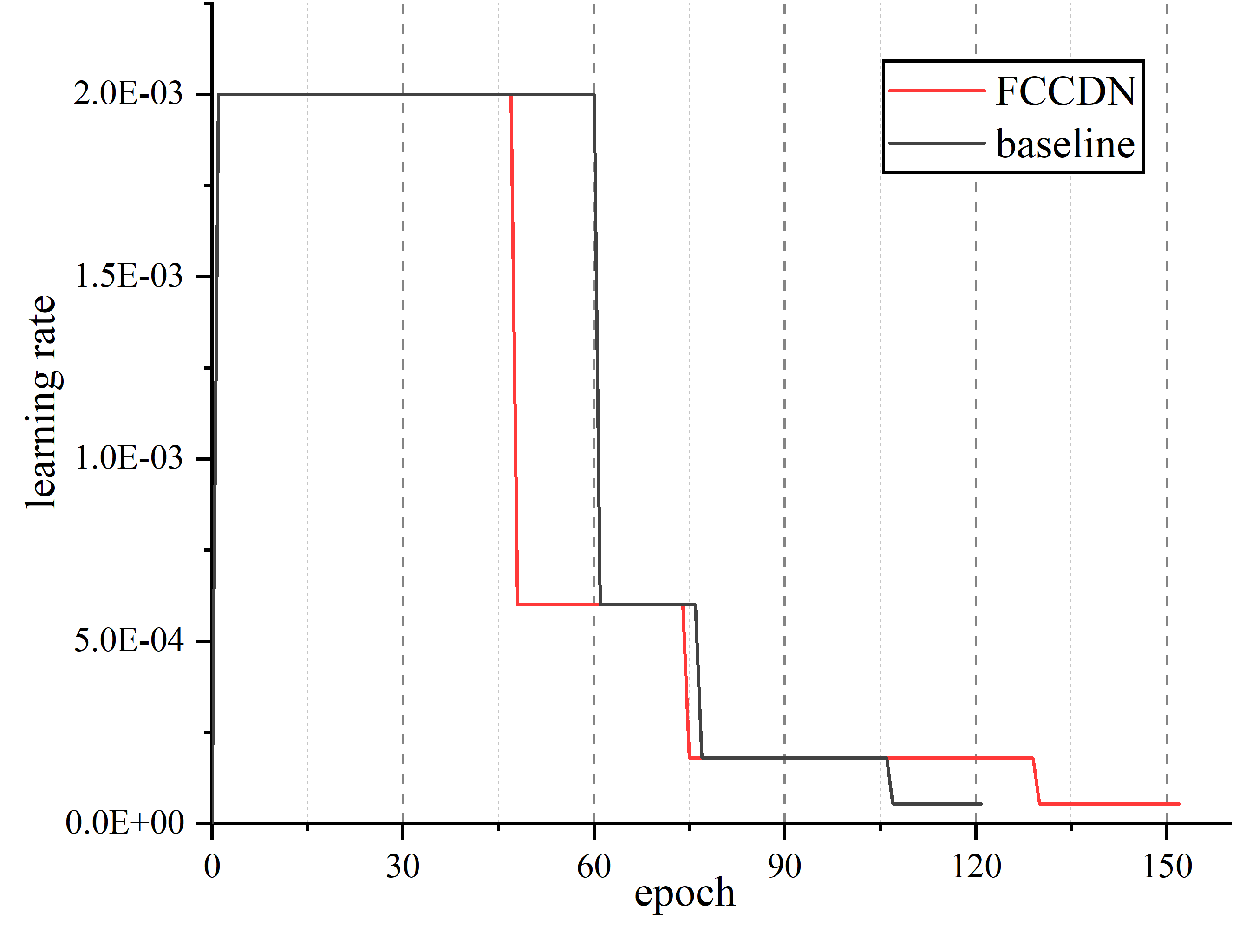}
\caption{The performance of the baseline and FCCDN on the validation sets. The top row corresponds to the LEVIR-CD dataset. The bottom row corresponds to the WHU dataset. We plot the curves of FCCDN in red and the curves of the baseline in gray. We highlight the best accuracy on the validation sets with green triangles on the curves.}
\label{fig:training}
\end{figure}

We plot the performance of the baseline (FCS) and FCCDN on the validation sets in Figure \ref{fig:training}. We show the curves of FCCDN in red and the curves of the baseline in gray. From Figure \ref{fig:training}, we can learn that FCCDN can achieve good performance after no more than 220 epochs. To further validate our models on the test sets, we save the weights with the highest validation accuracy as the checkpoints for testing.

\subsubsection{Inference}
\label{S:4.2.3}

We use a very simple inference methodology in the testing stage. On the LEVIR-CD dataset, we feed the original testing slices (sizes of 1024 × 1024) into the networks. On the WHU dataset, the slices (sizes of 256 × 256) are already prepared when splitting the dataset. Then, the slices are normalized with the same mean value and standard deviation used in the training stage. The checkpoints with the highest F1 scores on the validation sets are used for testing. The change masks are generated in the way described by Equation \ref{eq:get_pred}. We do not use test time augmentation (TTA) or multimodel ensembles.

\subsection{Evaluation Metrics}
\label{S:4.3}

In this paper, we use the intersection over union (IoU) and F1 score as the evaluation metrics. The two metrics are commonly used to quantify the performance of the change detection task. The values of IoU and F1 range from 0 to 1, and the higher the value is, the better the performance. The IoU and F1 score are calculated as:

\begin{equation}
IoU=\frac{TP}{TP+FP+FN}
\label{eq:iou}
\end{equation}
\begin{equation}
F1=2*\frac{precision*recall}{precision+recall}
\label{eq:f1}
\end{equation}

\noindent
The precision is calculated as:

\begin{equation}
precision=\frac{TP}{TP+FP}
\label{eq:precision}
\end{equation}

\noindent
The recall is calculated as:

\begin{equation}
recall=\frac{TP}{TP+FN}
\label{eq:recall}
\end{equation}

\noindent
TP denotes  true positive, FP denotes  false positive, and FN means false negative.

\subsection{Results}
\label{S:4.4}

\subsubsection{Ablation study}
\label{S:4.4.1}

In this subsection, we present the ablation study of FCCDN on the LEVIR-CD dataset. As is shown in Table \ref{Table:ablation_study}, we compare each architecture mainly with two evaluation metrics: the IoU and F1 score. The formulas of the above metrics are described in Section \ref{S:4.3}. We test each architecture three times and present the average performance in Table \ref{Table:ablation_study}.

We start the ablation study by composing two backbone architectures: the FCS network and DED network. FCS achieves the lowest IoU (0.8380) and F1 score (0.9119) in the ablation study. Although the precision of FCS is high (0.9301), its recall is extremely low (0.8944), which means FCS misses many changes. Compared with FCS, DED achieves great improvements (1\% in terms of the IoU and 0.6\% in terms of the F1 score). The performance of FCS and DED further proves our argument in Section \ref{S:3.2}.

We validate the NL-FPN by adding it to the center of FCS and DED. The performance is shown in the second line and the 6th line of Table \ref{Table:ablation_study}. FCS + NL-FPN achieves an IoU of 0.8430 and an F1 score of 0.9148. DED + NL-FPN achieves an IoU of 0.8500 and an F1 score of 0.9189. Notably, both FCS and DED benefit from NL-FPN.

We validate DFM by using it as the bitemporal feature fusion module in our architectures. The performance results displayed in the third line and the 7th line of Table \ref{Table:ablation_study} demonstrate that DFM can significantly improve the accuracy of change detection networks. The combination of FCS and DFM can boost the F1 score from 0.9119 to 0.9161. The combination of DED and DFM can boost the F1 score from 0.9176 to 0.9192.

For the SSL-based feature constraint strategy, we only test it with the DED backbone. By comparing the accuracy in the 5th line and the 8th line of Table \ref{Table:ablation_study}, we can see that the SSL-based strategy works as expected. It improves the IoU from 0.8478 to 0.8530 and improves the F1 score from 0.9176 to 0.9206.

We also try other combinations of the proposed architectures and show the experimental results in Table \ref{Table:ablation_study}. All the combinations can achieve distinct accuracy improvements. The combination of DED + NL-FPN + DFM + SSL achieves the best performance with an IoU of 0.8565 and an F1 score of 0.9227.

\begin{table}[ht]
\centering
\begin{tabular}{l c c c c c c c}
\hline
\textbf{} & {\footnotesize \textbf{NL-FPN}} & {\footnotesize \textbf{DFM}} & {\footnotesize \textbf{SSL}} & {\footnotesize \textbf{precision(\%)}} & {\footnotesize \textbf{recall(\%)}} & {\footnotesize \textbf{IoU(\%)}} & {\footnotesize \textbf{F1(\%)}} \\
\hline
{\footnotesize FCS} & & & $-$ & \textbf{93.01} & 89.44 & 83.80 ({\footnotesize ±0.1}) & 91.19 ({\footnotesize ±0.06}) \\
{\footnotesize FCS} & \checkmark & & $-$ & 92.76 & 86.90 & 84.30 ({\footnotesize ±0.26}) & 91.48 ({\footnotesize ±0.15}) \\
{\footnotesize FCS} & & \checkmark & $-$ & 92.78 & 87.06 & 84.53 ({\footnotesize ±0.12}) & 91.61 ({\footnotesize ±0.07}) \\
{\footnotesize FCS} & \checkmark & \checkmark & $-$ & 92.91 & 90.76 & 84.87 ({\footnotesize ±0.23}) & 91.82 ({\footnotesize ±0.14}) \\
{\footnotesize DED} & & & & 92.88 & 90.67 & 84.78 ({\footnotesize ±0.2}) & 91.76 ({\footnotesize ±0.12}) \\
{\footnotesize DED} & \checkmark & & & 92.66 & 91.15 & 85.00 ({\footnotesize ±0.17}) & 91.89 ({\footnotesize ±0.1}) \\
{\footnotesize DED} & & \checkmark & & 92.81 & 91.04 & 85.03 ({\footnotesize ±0.1}) & 91.92 ({\footnotesize ±0.05}) \\
{\footnotesize DED} & & & \checkmark & 92.97 & 91.20 & 85.30 ({\footnotesize ±0.11}) & 92.06 ({\footnotesize ±0.06}) \\
{\footnotesize DED} & \checkmark & \checkmark & & 92.45 & 91.75 & 85.35 ({\footnotesize ±0.12}) & 92.10 ({\footnotesize ±0.07}) \\
{\footnotesize DED} & \checkmark & & \checkmark & 92.31 & 91.73 & 84.98 ({\footnotesize ±0.3}) & 92.01 ({\footnotesize ±0.12}) \\
{\footnotesize DED} & & \checkmark & \checkmark & 92.81 & 91.47 & 85.42 ({\footnotesize ±0.2}) & 92.14 ({\footnotesize ±0.11}) \\
{\footnotesize DED} & \checkmark & \checkmark & \checkmark & 92.95 & \textbf{91.61} & \textbf{85.65}{\footnotesize (±0.05)} & \textbf{92.27} {\footnotesize (±0.02)} \\
\hline
\end{tabular}
\caption{The ablation study on the LEVIR-CD building change detection dataset. Values in bold font are the best. The symbol $-$ stands for no such combination. We present the performance shift with ±.}
\label{Table:ablation_study}
\end{table}

\subsubsection{LEVIR-CD dataset}
\label{S:4.4.2}

In this subsection, we present the comparison results of FCCDN and other change detection architectures on the LEVIR-CD dataset. The methods used for comparison are recently published methods that were tested on the LEVIR-CD dataset.

Most of the architectures are validated on the same test set proposed by \cite{chen2020spatial}, except for DDCNN \cite{peng2020optical}. Although DDCNN used a different data splitting approach, our method outperforms it with respect to the F1 score by 2\%. We list the comparison results in Table \ref{Table:levir_acc}. As shown in the table, our network architecture outperforms all the competitors and achieves new state-of-the-art results on the LEVIR-CD dataset.

\begin{table}[ht]
\centering
\begin{tabular}{l c c c c}
\hline
\textbf{} & \textbf{precision(\%)} & \textbf{recall(\%)} & \textbf{IoU(\%)} & \textbf{F1(\%)} \\
\hline
STANet \cite{chen2020spatial} & 83.80 & 91.00 & - & 87.30 \\
BiT S4 \cite{chen2021efficient} & 89.24 & 89.37 & 80.68 & 89.31 \\
DDCNN \cite{peng2020optical} & 91.85 & 88.69 & 82,21 & 90.24 \\
FracTAL ResNet \cite{diakogiannis2020looking} & 93.60 & 89.38 & 84.23 & 91.44 \\
CEECNetV1 \cite{diakogiannis2020looking} & 93.73 & 89.93 & 84.82 & 91.79 \\
CEECNetV2 \cite{diakogiannis2020looking} & \textbf{93.81} & 89.92 & 84.89  & 91.83 \\
\textbf{FCCDN (256)} & 92.96 & \textbf{91.55} & 85.62 & 92.25 \\
\textbf{FCCDN (512)} & 93.07 & 91.52 & \textbf{85.69} & \textbf{92.29} \\
\hline
\end{tabular}
\caption{The comparison on the LEVIR-CD change detection dataset. Values in bold font are the best.}
\label{Table:levir_acc}
\end{table}

\begin{figure}[H]
\centering
\subfigure[]{
\hspace{-0.5cm}
\includegraphics[width=0.16\linewidth]{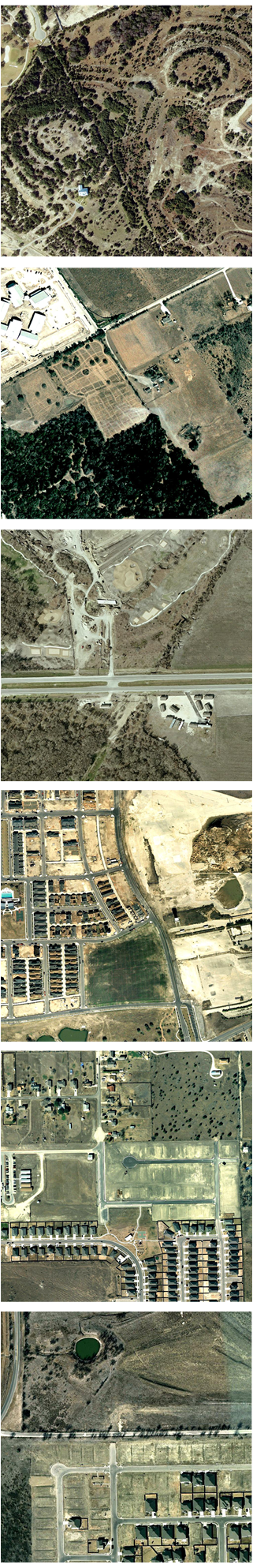}
\label{fig:levir_a}
}
\subfigure[]{
\hspace{-0.5cm}
\includegraphics[width=0.16\linewidth]{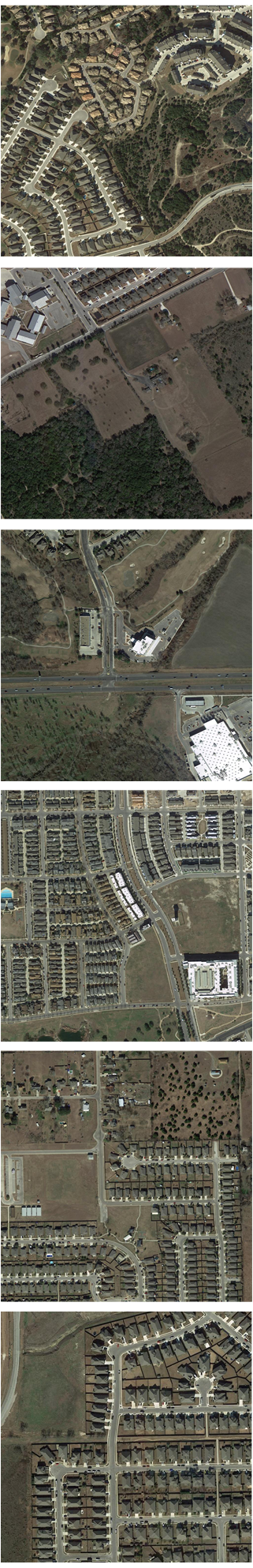}
\label{fig:levir_b}
}
\subfigure[]{
\hspace{-0.5cm}
\includegraphics[width=0.16\linewidth]{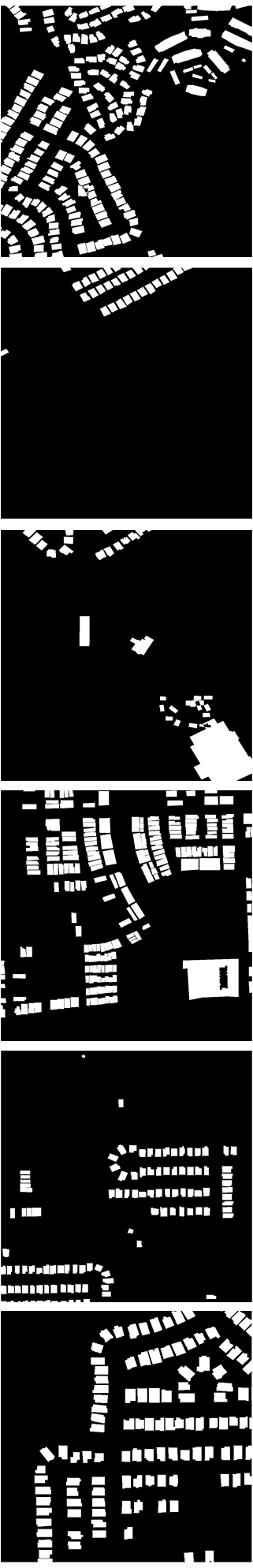}
\label{fig:levir_c}
}
\subfigure[]{
\hspace{-0.5cm}
\includegraphics[width=0.16\linewidth]{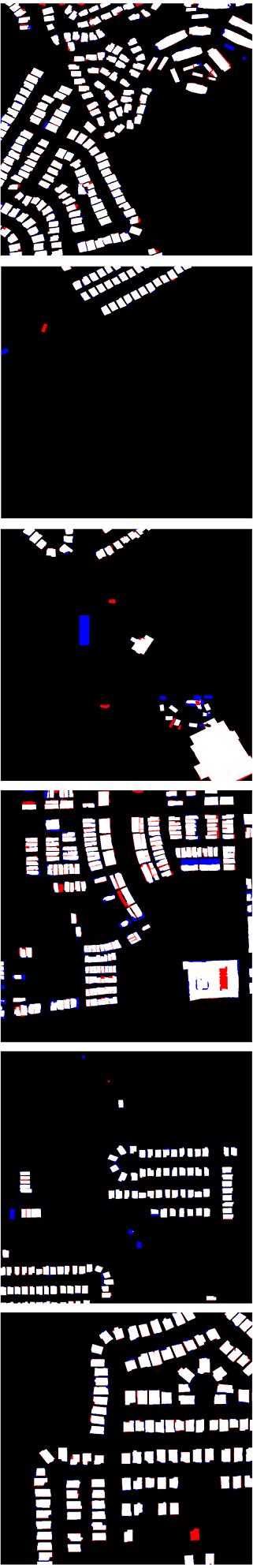}
\label{fig:levir_d}
}
\subfigure[]{
\hspace{-0.5cm}
\includegraphics[width=0.16\linewidth]{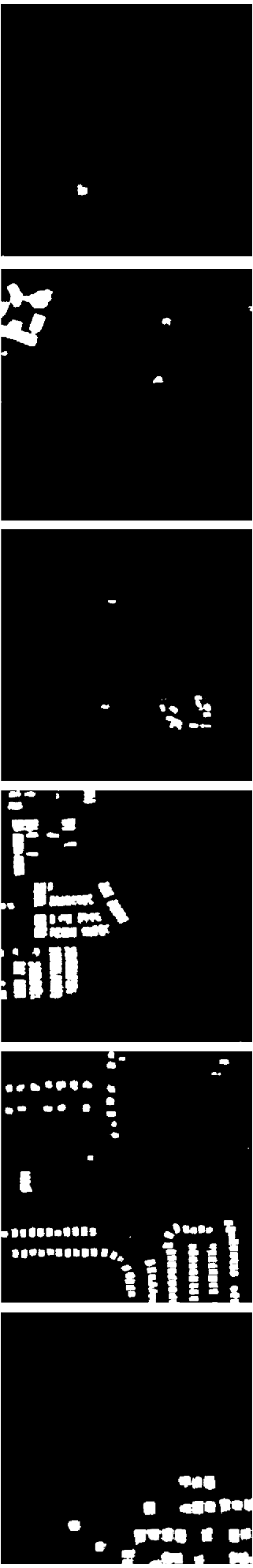}
\label{fig:levir_e}
}
\subfigure[]{
\hspace{-0.5cm}
\includegraphics[width=0.16\linewidth]{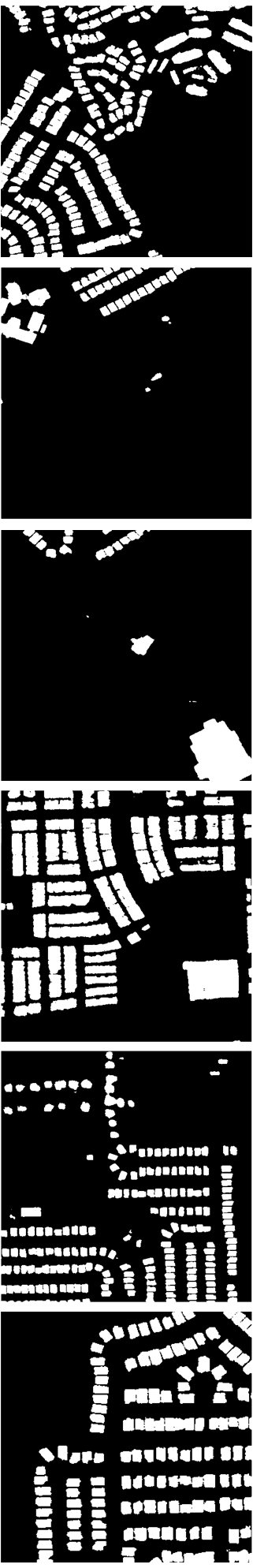}
\label{fig:levir_f}
}
\caption{Performance on the LEVIR-CD building change detection dataset. (a) Pretemporal images. (b) Posttemporal images. (c) Ground truth images. (d) Results of FCCDN. (e) Building segmentation results of T1. (f) Building segmentation results of T2. We plot the false positives in red and the false negatives in blue.}
\label{fig:levir_show}
\end{figure}

We present several inference results on the test set in Figure \ref{fig:levir_show}. Since almost all existing methods are validated on the same test set from this dataset, the values in Table \ref{Table:levir_acc} represent the performance of the existing methods. We only plot our results in this figure to show the inference details. The first three columns of the figure are the bitemporal images and ground truths. In the fourth column, we present the results of FCCDN. To better identify the difference between the labels and change masks, we plot the change masks in four colors: white denotes changed areas that are correctly identified (true positive), black denotes unchanged areas that are correctly identified (true negative), red denotes unchanged areas that are wrongly identified as changed areas (false positive), blue denotes changed areas that are missed (false negative). In addition to the change masks generated by FCCDN, we also plot the bitemporal semantic segmentation results of buildings in the fifth and sixth columns, which are generated in an unsupervised way.

As shown in Figure \ref{fig:levir_show}, FCCDN performs excellently on the test set. Almost all the changes are accurately found. We need to note that there are still some significant differences at the edges of the buildings, especially in the first row and the fourth row. These differences are mainly caused by the shadows of buildings. It is difficult to identify the exact boundary since some edges are hidden in the shadows. 

In the field of observing landcover information, we not only care about the changing information but also care about the information of target objects in bitemporal images. With the above information, we can further obtain advanced results. Therefore, the pixel category in bitemporal images is vital for the application of change detection. As shown in the last two columns of Figure \ref{fig:levir_show}, we output accurate bitemporal semantic segmentation results of the buildings in an unsupervised way. This is the first study to generate bitemporal semantic segmentation results only with change labels; existing methods need extra segmentation labels. In conclusion, our architecture can save much labeling work in real applications. 

\vspace{-0.45cm}

\subsubsection{WHU dataset}
\label{S:4.4.3}

We present the comparison results of FCCDN and other change detection architectures on the WHU dataset in this subsection. The comparison is mainly based on the work of \cite{peng2020optical}, who tested many change detection methods with their splitting approach. Therefore, we use the values of the comparison methods reported in their work. We also add the experimental result of \cite{chen2021efficient}, who split their dataset in the same way as in \cite{peng2020optical}. We list the comparison results in Table \ref{Table:whu_acc}. From the table, we can learn that FCCDN outperforms all other architectures with a remarkable advantage. We achieve the highest IoU (0.8820) and F1 score (0.9373) on this dataset.

\begin{table}[ht]
\centering
\begin{tabular}{l c c c c}
\hline
\textbf{} & \textbf{precision(\%)} & \textbf{recall(\%)} & \textbf{IoU(\%)} & \textbf{F1(\%)} \\
\hline
FC-EF \cite{daudt2018fully} & 78.86 & 78.64 & 64.94 & 78.75 \\
FC-Sima-diff \cite{daudt2018fully} & 84.73 &	87.31 &	75.44 &	86.00 \\
FC-Sima-conc \cite{daudt2018fully} & 78.86 & 78.64 & 64.94 & 78.75 \\
BiDataNet \cite{papadomanolaki2019detecting} & 86.75 & 90.60 & 79.59 & 88.63 \\
BiT S4 \cite{chen2021efficient} & 86.64 & 81.48 & 72.39 & 83.98 \\
CDNet \cite{alcantarilla2018street} & 91.75 & 86.89 & 80.60 & 89.62 \\
UNet++\_MSOF \cite{peng2019end} & 91.96 & 89.40 & 82.92 & 90.66 \\
DASNet \cite{chen2020dasnet} & 88.23 & 84.62 & 76.04 & 86.39 \\
DDCNN \cite{peng2020optical} & 93.71 & 89.12 & 84.09 & 91.36 \\
IFN \cite{zhang2020deeply} & 91.44 & 89.75 & 82.79 & 90.59 \\
\textbf{FCCDN (ours)} & \textbf{96.39} & \textbf{91.24} & \textbf{88.20} & \textbf{93.73} \\
\hline
\end{tabular}
\caption{The comparison results on the WHU building change detection dataset. The values in bold font are the best.}
\label{Table:whu_acc}
\end{table}

\begin{figure}[H]
\centering
\subfigure[]{
\hspace{-0.5cm}
\includegraphics[width=0.14\linewidth]{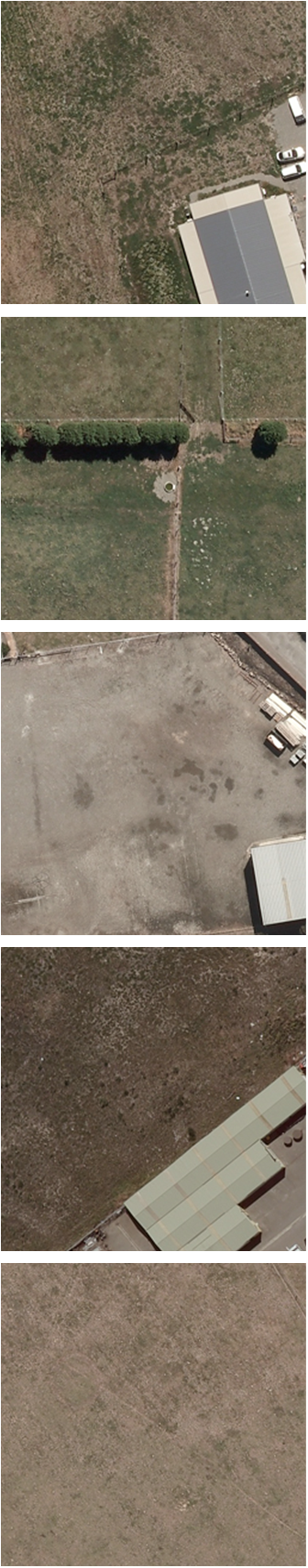}
\label{fig:whu_a}
}
\subfigure[]{
\hspace{-0.5cm}
\includegraphics[width=0.14\linewidth]{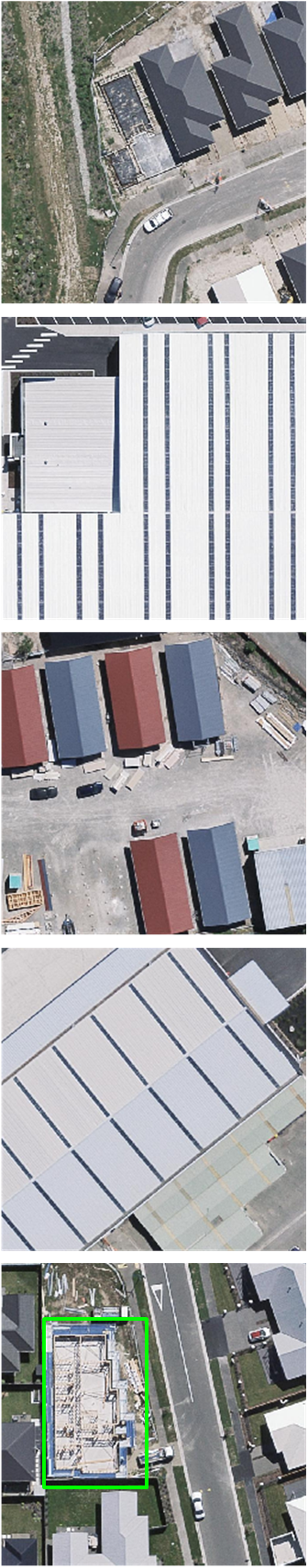}
\label{fig:whu_b}
}
\subfigure[]{
\hspace{-0.5cm}
\includegraphics[width=0.14\linewidth]{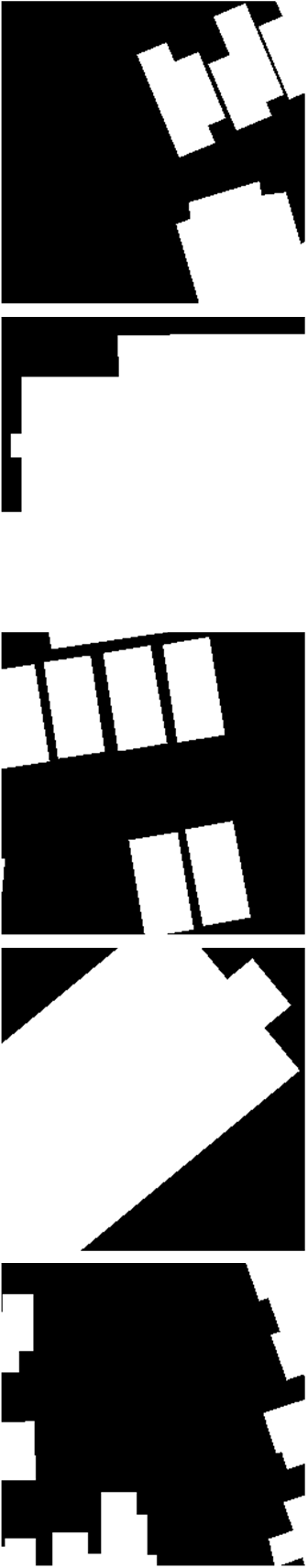}
\label{fig:whu_c}
}
\subfigure[]{
\hspace{-0.5cm}
\includegraphics[width=0.14\linewidth]{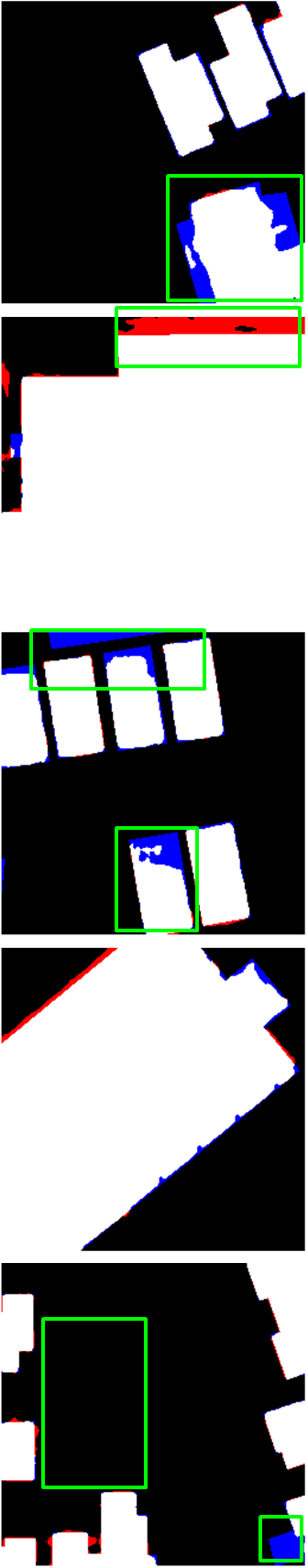}
\label{fig:whu_d}
}
\subfigure[]{
\hspace{-0.5cm}
\includegraphics[width=0.14\linewidth]{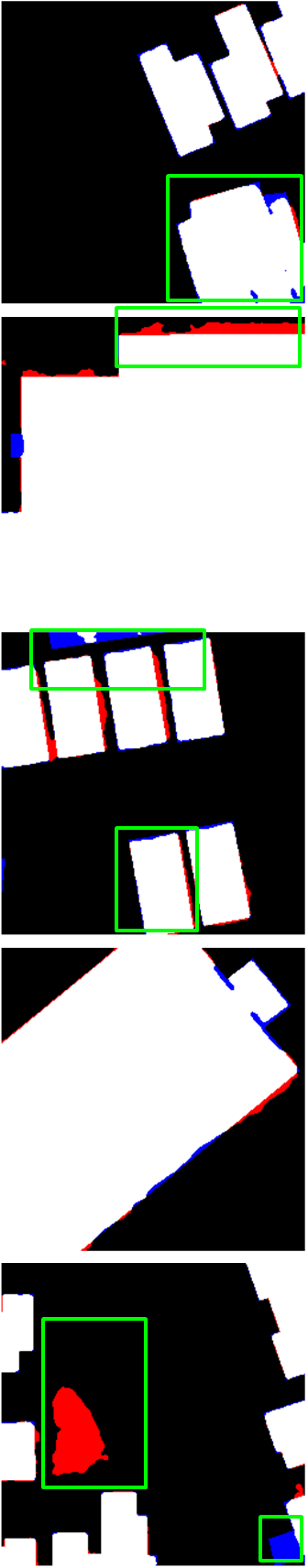}
\label{fig:whu_e}
}
\subfigure[]{
\hspace{-0.5cm}
\includegraphics[width=0.14\linewidth]{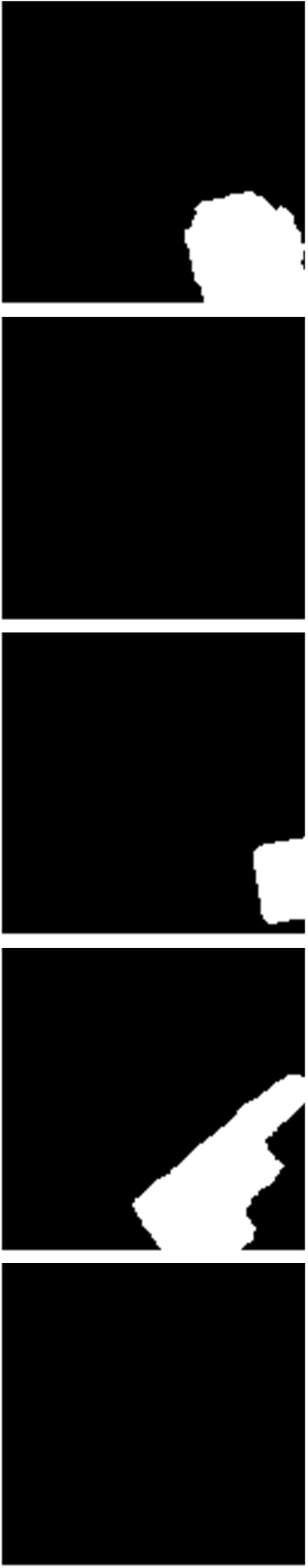}
\label{fig:whu_f}
}
\subfigure[]{
\hspace{-0.5cm}
\includegraphics[width=0.14\linewidth]{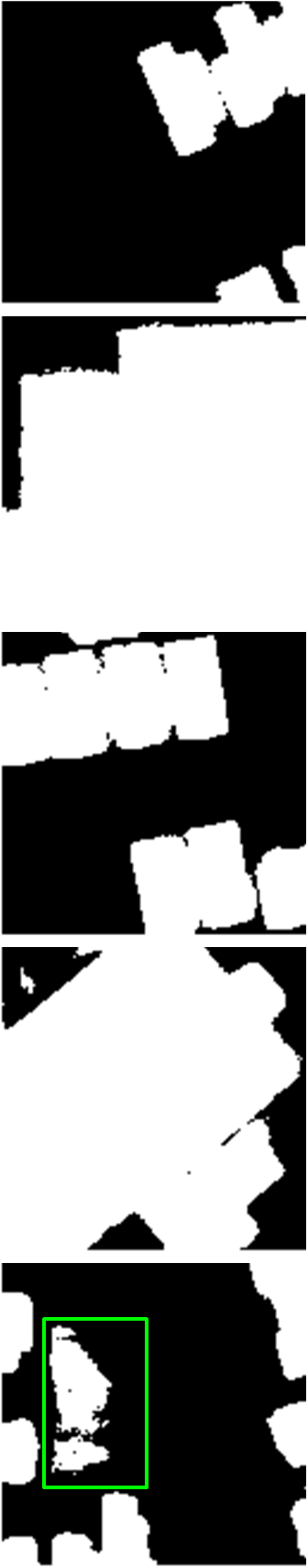}
\label{fig:whu_g}
}
\caption{Performance on the WHU building change detection dataset. (a) Pretemporal images. (b) Posttemporal images. (c) Ground truth images. (d) Results of DDCNN. (e) Results of FCCDN. (f) Pretemporal building segmentation results. (g) Posttemporal building segmentation results. We highlight interesting regions with green rectangles. We plot the false positives in red and the false negatives in blue.}
\label{fig:whu_show}
\end{figure}

We plot several inference results of the test set in Figure \ref{fig:whu_show}. In the first three columns, we present the bitemporal images and ground truths. The fourth column and the fifth column are the change masks of DDCNN and FCCDN. We also show the bitemporal semantic segmentation results in the last two columns. The distribution of the slices for visualization is shown in Figure \ref{fig:whu_show}. Each green rectangle indicates a testing chip of size 256 × 256.

\vspace{-0.45cm}

To better identify the difference between the labels and change masks, we plot the change masks with four colors in the same way as mentioned before. We further highlight some critical areas with green rectangles. As shown in the figure, FCCDN achieves excellent performance on the test set and outperforms the comparison methods. Most of the changes are well recognized except in several areas. At the tops of the second and third rows, both our results and the comparison results contain some incorrect decisions. This is understandable because it is difficult to identify buildings at the edge of images. Inferences with larger slices can perform better. At the center of the last row, there are pseudochanges in the result of FCCDN. However, since the building in the posttemporal image is under construction, building changes do exist in this area. At the bottom right of the last row, both FCCDN and the comparison architecture miss a small change. This seems to be a common problem in this dataset and is also mentioned in \cite{diakogiannis2020looking}. 

\begin{figure}[ht]
\centering
\includegraphics[width=1.0\linewidth]{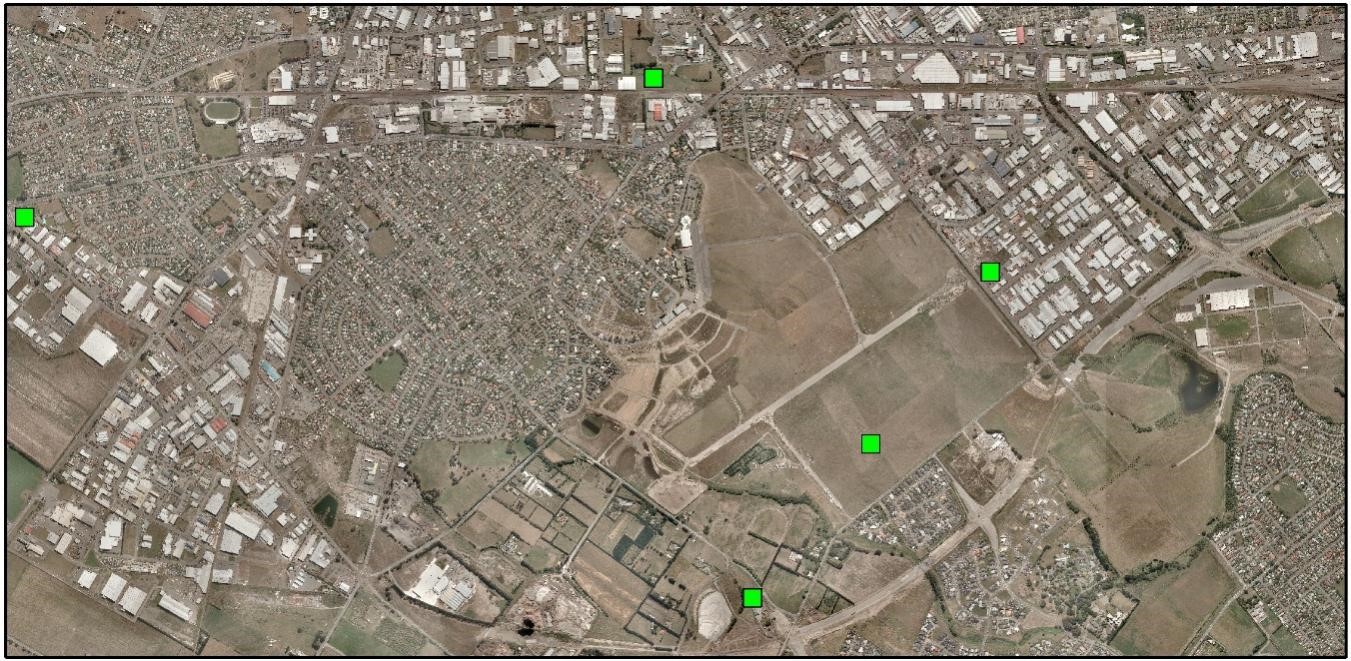}
\caption{Distribution of slices for visualization on the WHU dataset. The green rectangles represent the 256 × 256 slices we used for visualization.}
\label{fig:whu_test_pos}
\end{figure}

\subsubsection{efficiency test}
\label{S:4.4.4}

In this subsection, we present the efficiency of FCCDN in terms of four factors: parameters of the model (Params), the training time (Tt), the training batch size with 12 GB memory (Tb/12 GB), and the inference time with batch size = 1 (It). We evaluate our model efficiency on the WHU dataset. To better show the efficiency of FCCDN, we also test the efficiency of several existing change detection methods \cite{daudt2018fully,chen2020dasnet,alcantarilla2018street,peng2019end,peng2020optical,zhang2020deeply}.

Table \ref{Table:efficiency_test} shows the comparison results of the efficiency test. From the table, we can see that FCCDN achieves the best F1 score with competitive efficiency.

\begin{table}[ht]
\centering
\begin{tabular}{l c c c c c}
\hline
\textbf{} & \textbf{Params(Mb)} & \textbf{Tt(s)} & \textbf{It(s)} & \textbf{F1(\%)} \\
\hline
FC-EF \cite{daudt2018fully} & 5.15 & $\thicksim $140 & $\thicksim $35 & 90 & 78.75 \\
FC-Sima-diff \cite{daudt2018fully} & 5.15 & $\thicksim $140 & $\thicksim $35 & 65 & 86.00 \\
FC-Sima-conc \cite{daudt2018fully} & 5.9 & $\thicksim $140 & $\thicksim $35 & 60 & 83.47 \\
BiDataNet \cite{papadomanolaki2019detecting} & 192 & $\thicksim $140 & $\thicksim $45 & 40 & 88.63 \\
BiT S4 \cite{chen2021efficient} & 7.75 & $\thicksim $145 & $\thicksim $40 & 45 & 89.62 \\
CDNet \cite{alcantarilla2018street} & 7.75 & $\thicksim $145 & $\thicksim $40 & 45 & 89.62 \\
UNet++\_MSOF \cite{peng2019end} & 34.6 & $\thicksim $170 & $\thicksim $40 & 14 & 90.66 \\
DDCNN \cite{peng2020optical} & 178 & $\thicksim $750 & $\thicksim $85 & 5 & 91.36 \\
IFN \cite{zhang2020deeply} & 137 & $\thicksim $255 & $\thicksim $75 & 14 & 90.59 \\
\textbf{FCCDN (ours)} & \textbf{24.2} & \textbf{$\thicksim $150} & \textbf{$\thicksim $60} & \textbf{28} & \textbf{93.73} \\
\hline
\end{tabular}
\caption{The comparison on WHU building change detection dataset. Values in bold font are the best.}
\label{Table:efficiency_test}
\end{table}

\section{Discussion}
\label{S:5}

\subsection{SSL-based strategy for multi-class change tasks}
\label{S:5.1}

\begin{figure}[H]
\centering
\subfigure[]{
\hspace{-0.5cm}
\includegraphics[width=0.8\linewidth]{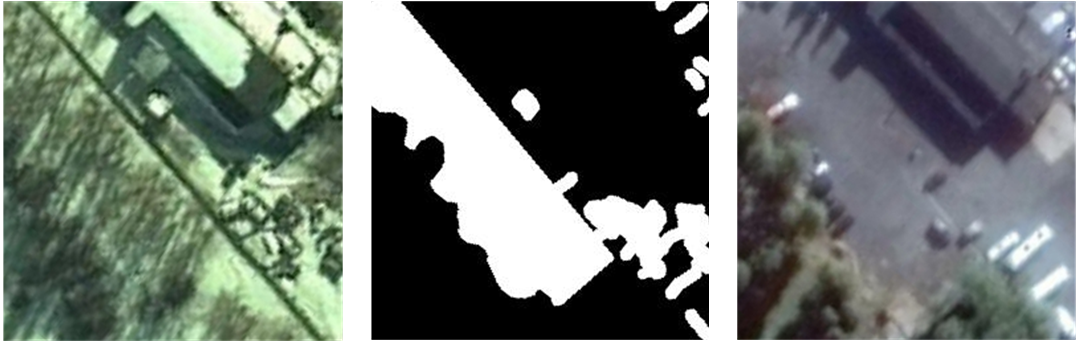}
\label{fig:vhr}
}
\subfigure[]{
\hspace{-0.5cm}
\includegraphics[width=0.8\linewidth]{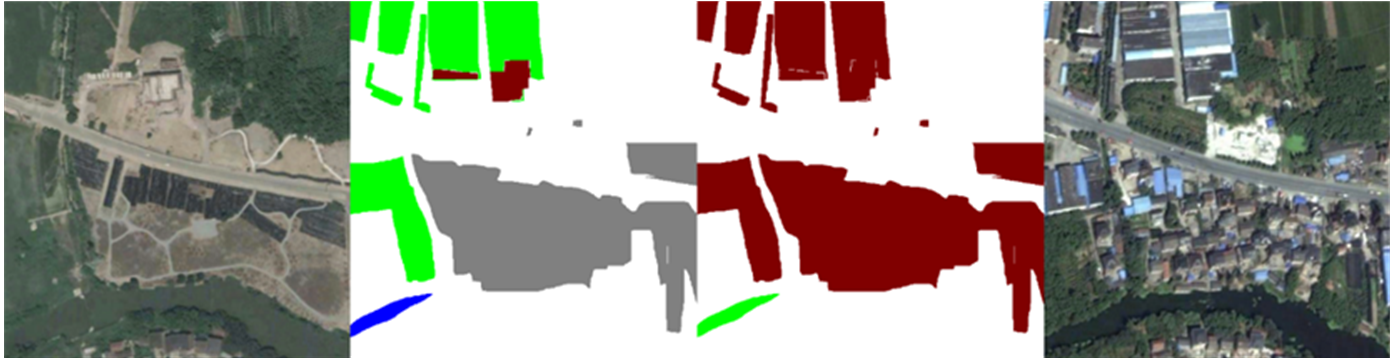}
\label{fig:second}
}
\subfigure[]{
\hspace{-0.5cm}
\includegraphics[width=0.8\linewidth]{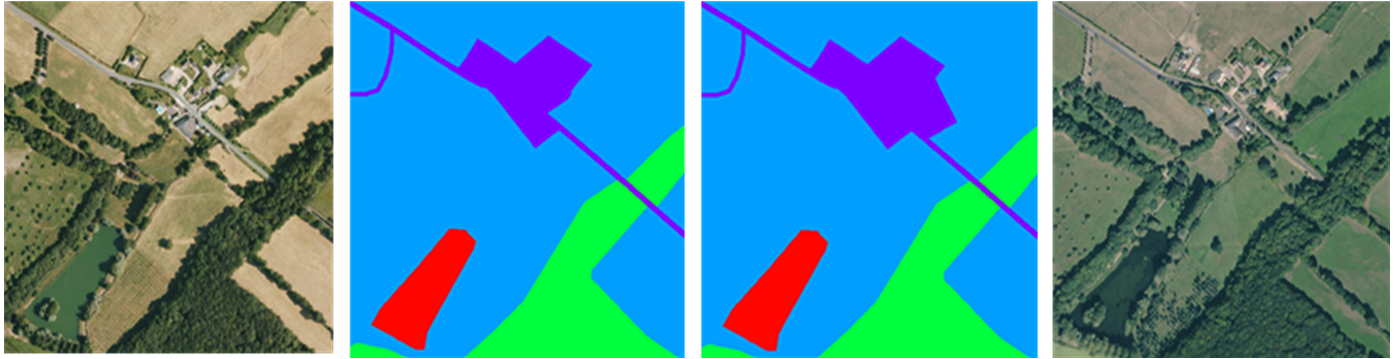}
\label{fig:hrscd}
}
\caption{Examples of multiclass change detection datasets. (a) Samples of the season-varying dataset. From left to right are the pretemporal image, change label, and posttemporal image. (b) Samples of the SECOND dataset. From left to right are the pretemporal image, pretemporal changed area semantic segmentation label, posttemporal changed area semantic segmentation label, and posttemporal image. (c) Samples of the HRSCD dataset. From left to right are the pretemporal image, pretemporal semantic segmentation label, posttemporal semantic segmentation label, the posttemporal image.}
\label{fig:multi-class_datasets}
\end{figure}

In this subsection, we discuss the application of the proposed SSL-based feature constraint strategy to multiclass change detection tasks. The discussion is addresses three popular multiclass change detection datasets: 1) the season-varying dataset \cite{lebedev2018change}, which is used to identify multiclass changes with binary change labels (Figure \ref{fig:vhr}); 2) SECOND \cite{yang2020asymmetric}, which is used to identify multiclass changes with multiclass semantic segmentation labels in the bitemporal changed area (Figure \ref{fig:second}); 2) HRSCD \cite{daudt2019multitask}, which is used to identify multiclass changes with multiclass semantic segmentation labels in each temporal image (Figure \ref{fig:hrscd}).

\vspace{-0.45cm}

\subsubsection{Season-varying dataset}
\label{S:5.1.1}

The season-varying dataset is a multiclass change detection dataset with optical satellite images obtained by GE. The resolution of the dataset varies from 3 cm to 100 cm. As shown in Figure \ref{fig:vhr}, seasonal changes are labeled with white color, and the rest are unchanged objects. The author of the dataset released a standard splitting plan, which consists of 10000 training slices, 3000 validation slices, and 3000 test slices. Each slice is an image with a size of 256 × 256.

Since the SSL-based feature constraint strategy relies on the supervision of the segmentation branches, it is essential to know the categories of the changed objects. The pseudolabels can be obtained only by acknowledging the class of objects in the bitemporal images. For the season-varying dataset, there are no clear categories for the changed objects. In this case, our SSL-based strategy will be simplified to a contrastive loss function. The contrastive loss is used to pull closer unchanged bitemporal features and push changed bitemporal features away. The loss function can be formulated as:

\begin{equation}
l_{aux}=MSE\left ( S_{1,i}, S_{2,i}|i\in U\right ) - MSE\left ( S_{1,i}, S_{2,i}|i\in C\right )
\label{eq:contra_loss}
\end{equation}

where $l_{\rm aux}$ denotes the contrastive loss, $U$ denotes the unchanged pixels, $C$ denotes the changed pixels, $S_{\rm *}$ denotes the outputs of the auxiliary branches, and MSE denotes the mean squared error (squared L2 norm), which is calculated as:

\begin{equation}
l_{MSE}=\frac{1}{N}\sum_{n=1}^{N}\left ( x_{n}-y_{n} \right )^{2}
\label{eq:MSE}
\end{equation}

where N represents the total number of pixels in a label patch. The contrastive loss is added to the main loss in the following way:

\begin{equation}
loss=l_{change}+0.2*l_{aux}
\label{eq:vhr_loss}
\end{equation}

where $l_{\rm change}$ is the loss of the binary change mask and is calculated with the BCE loss + dice coefficient loss.

We test the contrastive loss strategy on the season-varying dataset by fitting it to the DED backbone (Figure \ref{fig:DED}). We add two additional branches to DED. The output of the two branches is fed into the auxiliary loss function. We use the same training details as the details shown in Section \ref{S:4.2}. The experimental result is shown in Table \ref{Table:vhr_acc}. From the table, we can see that the contrastive loss strategy can remarkably boost the IoU performance from 0.9081 to 0.9157 and the F1 score from 0.9519 to 0.9560.

\begin{table}[ht]
\centering
\begin{tabular}{l c c c c}
\hline
\textbf{} & \textbf{precision(\%)} & \textbf{recall(\%)} & \textbf{IoU(\%)} & \textbf{F1(\%)} \\
\hline
DED & 97.22 & 93.23 & 90.81 & 95.19 \\
\textbf{DED + ContraLoss} & \textbf{97.33} & \textbf{97.93} & \textbf{91.57} & \textbf{95.60} \\
\hline
\end{tabular}
\caption{Experimental result on the season-varying dataset. DED in the table represents the dual encode-decode backbone proposed in Section \ref{S:3.2}. ContraLoss represents the contrastive loss strategy.}
\label{Table:vhr_acc}
\end{table}

\begin{figure}[H]
\centering
\subfigure[]{
\hspace{-0.5cm}
\includegraphics[width=0.19\linewidth]{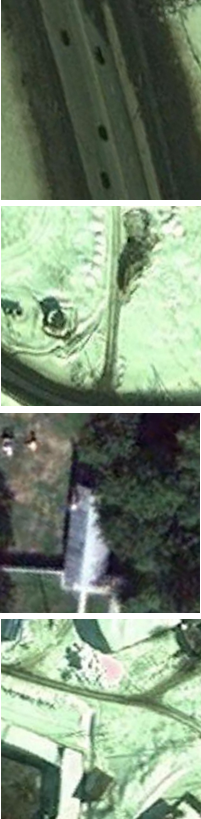}
\label{fig:vhr_a}
}
\subfigure[]{
\hspace{-0.5cm}
\includegraphics[width=0.19\linewidth]{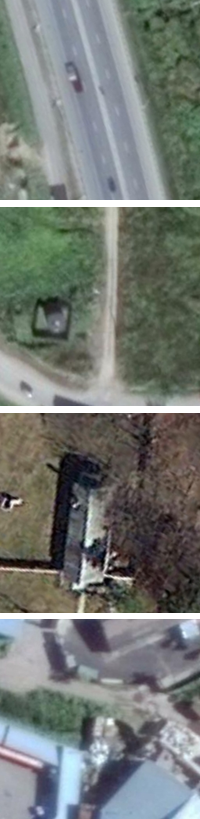}
\label{fig:vhr_b}
}
\subfigure[]{
\hspace{-0.5cm}
\includegraphics[width=0.19\linewidth]{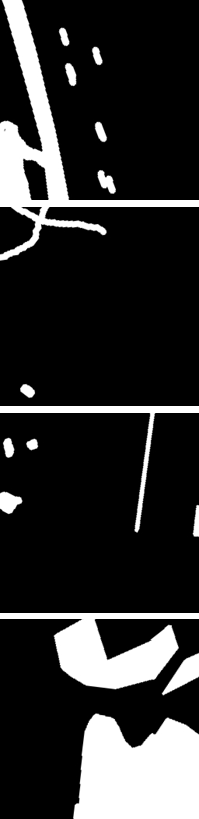}
\label{fig:vhr_c}
}
\subfigure[]{
\hspace{-0.5cm}
\includegraphics[width=0.19\linewidth]{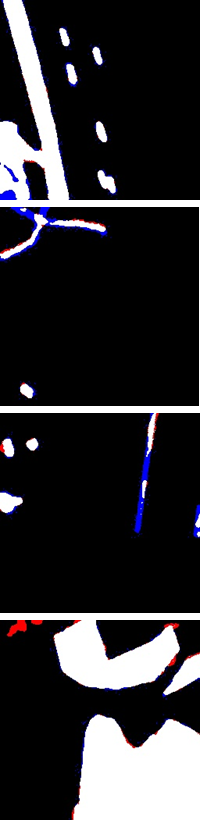}
\label{fig:vhr_d}
}
\subfigure[]{
\hspace{-0.5cm}
\includegraphics[width=0.19\linewidth]{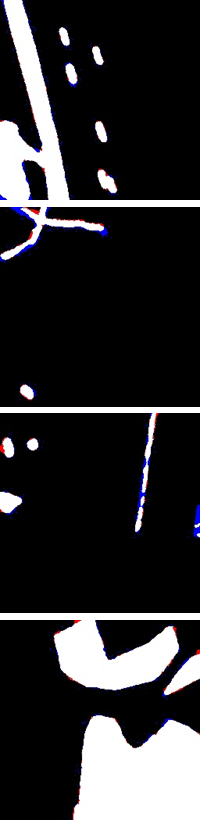}
\label{fig:vhr_e}
}
\caption{Comparison results on the season-varying dataset. (a) Pretemporal images. (b) Posttemporal images. (c) Ground truth images. (d) Results of DED. (e) Results of DED with the contrastive loss. We plot the false positives in red and the false negatives in blue.}
\label{fig:vhr_show}
\end{figure}

We also visualize several test results on the season-varying dataset in Figure \ref{fig:vhr_show}. The results show that the contrastive loss strategy can help the model better identify seasonal changes and suppress unconcerned changes.

\vspace{-0.45cm}

\subsubsection{SECOND dataset}
\label{S:5.1.2}

The SECOND dataset is a multiclass change detection dataset with six landcover classes, i.e., nonvegetated ground surface, trees, low vegetation, water, buildings, and playgrounds. Changed objects in the bitemporal images are annotated with the above categories, while the unchanged objects are labeled as background. Samples of the SECOND dataset are shown in Figure \ref{fig:second}. From left to right are the pretemporal image, pretemporal changed area semantic segmentation label, posttemporal changed area semantic segmentation label, and and posttemporal image. Changed objects are labeled with different values (visualized with different colors), and unchanged objects are masked out (visualized with white). The dataset contains 2968 samples. We randomly split the dataset into a training set, validation set, and testing set based on the ratio train:validation:test=8:1:1.

With the explicit categories of changed objects, we can further constrain feature learning by the additional semantic segmentation tasks on the bitemporal images. To this end, we add two semantic segmentation branches to the DED backbone. Each branch produces a semantic segmentation result. Since only bitemporal semantic segmentation labels in the changed area are available, we utilize a multiclass SSL strategy for the unchanged area. We first identify the unchanged area and the changed area with change detection labels. In the changed area, the output is supervised with a typical semantic segmentation task. In the unchanged area, the semantic segmentation result of one branch is used as the label of the other branch. The SSL-based loss function can be expressed as:

\begin{equation}
l_{u}=F\left ( S_{1,i}, P_{2,i}|i\in U\right ) + F\left ( S_{2,i}, P_{1,i}|i\in C\right )
\label{eq:ms_ssl}
\end{equation}

where $l_{\rm u}$ donates the SSL-based loss in the unchanged area, $U$ donates unchanged pixels, $S_{\rm *}$ donates the outputs of auxiliary branches, $P_{\rm *}$ donates the semantic segmentation results of auxiliary branches, $F$ donates Crossentropy loss + dice coefficient loss. The final loss can be calculated as:

\begin{equation}
loss=0.5*l_{change}+0.5*l_{c}+0.2*l_{u}
\label{eq:second_loss}
\end{equation}

where $l_{\rm change}$  represents the loss of the binary change mask, $l_{\rm c}$ represents the semantic segmentation loss in the changed area, and $l_{\rm u}$ represents the SSL-based loss. All the loss values are calculated with the cross-entropy loss + dice coefficient loss.

We validate the performance of the multiclass SSL strategy by applying it to ordinary DED, which has one change mask branch and two semantic segmentation branches for the changed pixels. Compared with ordinary DED, the SSL-based strategy introduces two additional semantic segmentation branches for the unchanged pixels. We use the mean IoU (mIoU) of the binary change mask and the separated kappa (Sek) coefficient \cite{yang2020asymmetric} of the multiclass change detection results as the evaluation metrics. The Sek coefficient is a modified kappa coefficient that alleviates the influence of label imbalance. Table \ref{Table:second_acc} shows the comparison results.

\begin{table}[ht]
\centering
\begin{tabular}{l c c}
\hline
\textbf{} & \textbf{mIoU (\%)} & \textbf{Sek(\%)} \\
\hline
DED & 70.34 & 19.17 \\
\textbf{DED + SSL} & \textbf{70.79} & \textbf{20.1} \\
\hline
\end{tabular}
\caption{Experimental results on the SECOND dataset. DED in the table represents the dual encode-decode backbone proposed in Section \ref{S:3.2}. SSL represents the multiclass SSL strategy.}
\label{Table:second_acc}
\end{table}

\begin{figure}[H]
\centering
\subfigure[]{
\hspace{-0.5cm}
\includegraphics[width=0.9\linewidth]{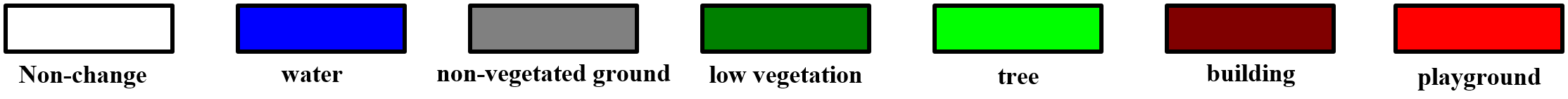}
\label{fig:SECOND_color}
}
\subfigure[]{
\hspace{-0.5cm}
\includegraphics[width=0.24\linewidth]{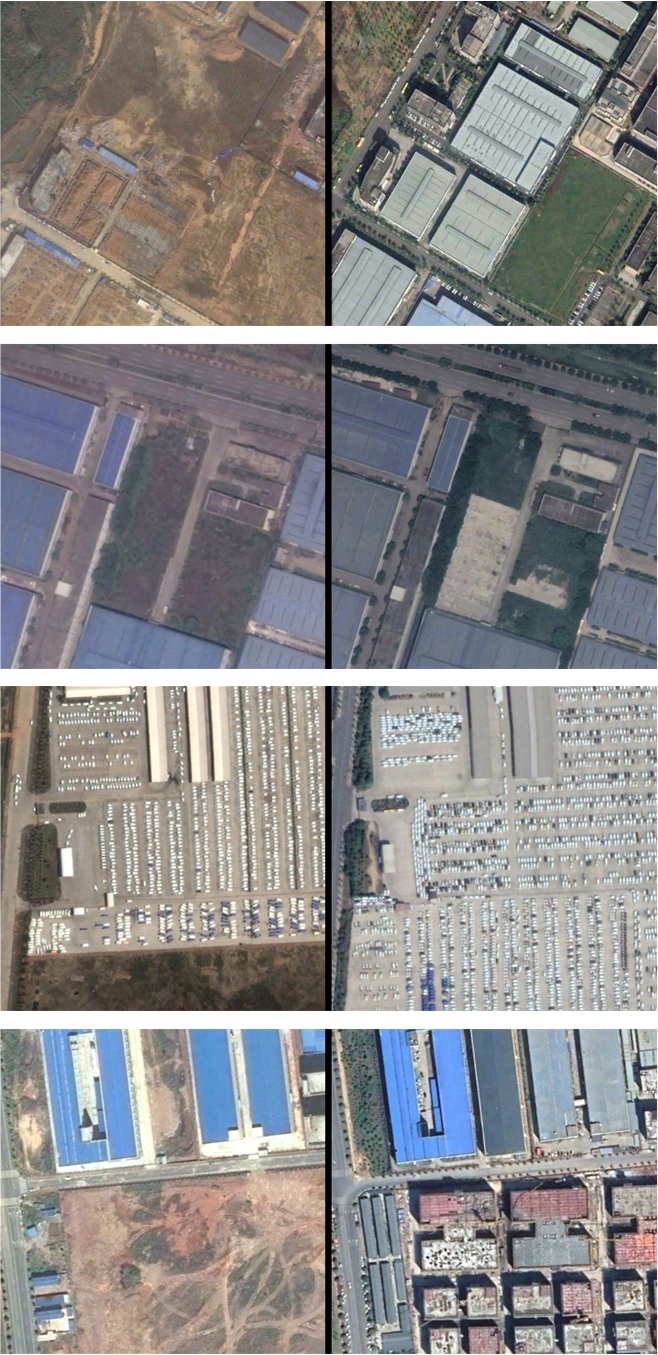}
\label{fig:SECOND_a}
}
\subfigure[]{
\hspace{-0.5cm}
\includegraphics[width=0.24\linewidth]{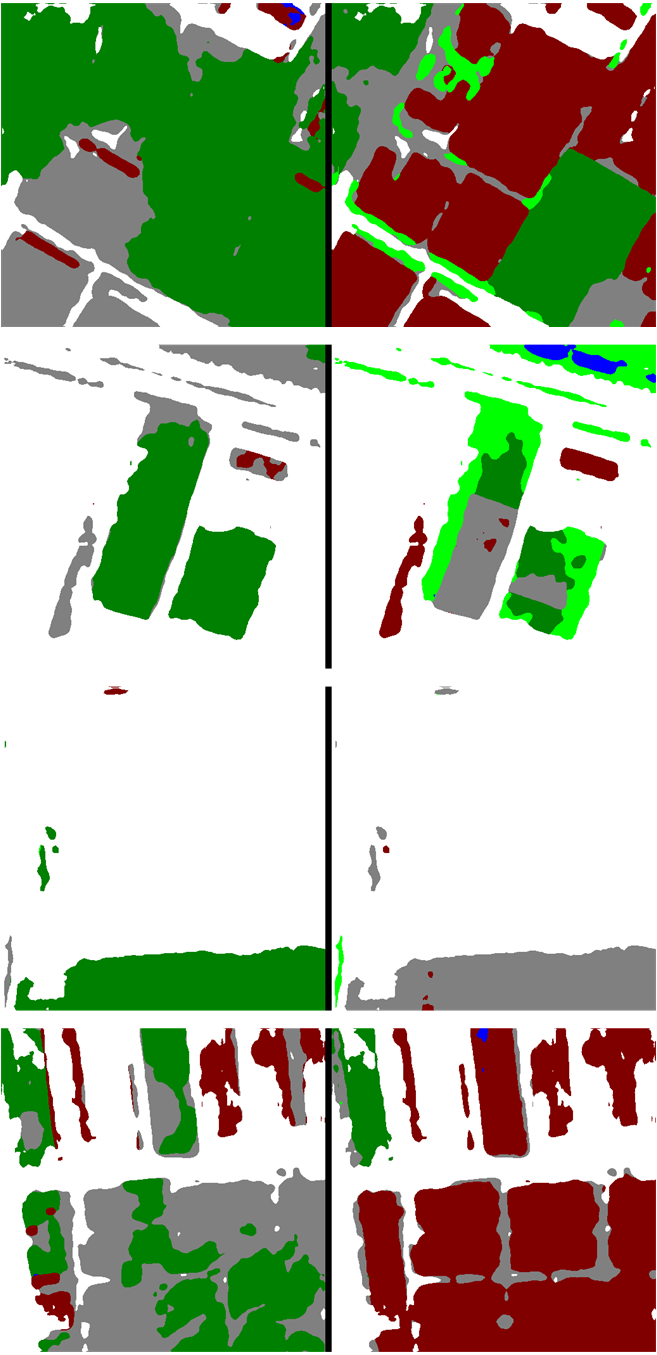}
\label{fig:SECOND_b}
}
\subfigure[]{
\hspace{-0.5cm}
\includegraphics[width=0.24\linewidth]{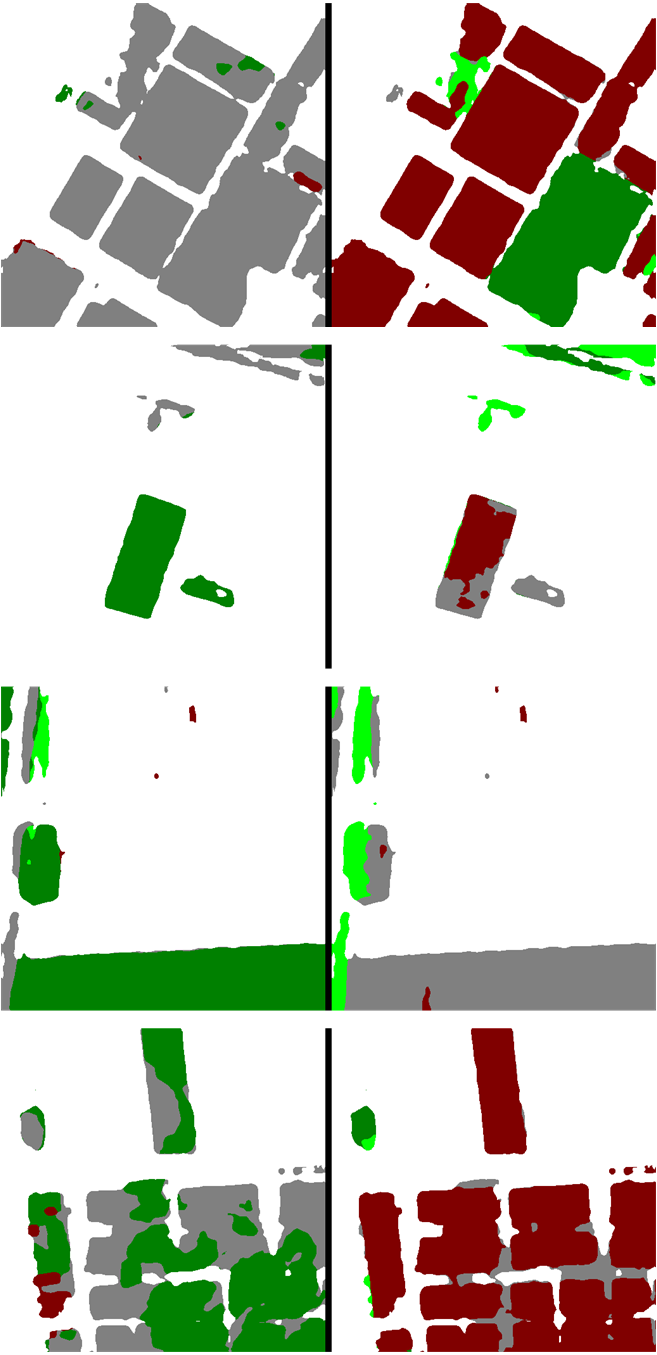}
\label{fig:SECOND_c}
}
\subfigure[]{
\hspace{-0.5cm}
\includegraphics[width=0.24\linewidth]{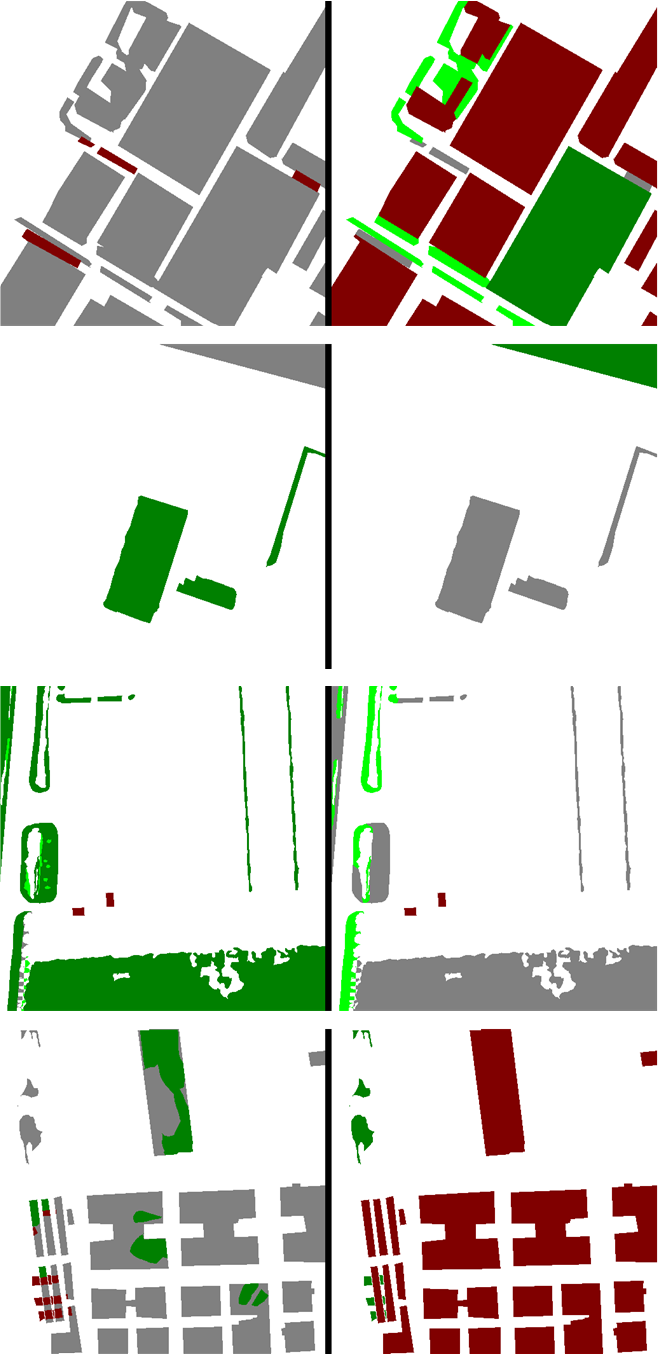}
\label{fig:SECOND_d}
}
\caption{Comparison results on the SECOND dataset. (a) Legend. (b) Bitemporal images. (c) Results of ordinary DED. (d) Results of DED with the SSL-based strategy. (e) Ground truth images. Changed objects are visualized with different colors and unchanged objects are visualized with white.}
\label{fig:SECOND_show}
\end{figure}

Figure \ref{fig:SECOND_show} shows several inference results on the testing set. From the figure, we can see that the model performs better with the supervision of pseudolabels in the unchanged area. Unconcerned changes are effectively suppressed in the final results.

\vspace{-0.45cm}

\subsubsection{HRSCD dataset}
\label{S:5.1.3}

The HRSCD dataset is a multiclass change detection dataset with bitemporal semantic segmentation labels. Objects in bitemporal images are divided into several semantic classes, i.e., no information, artificial surfaces, agricultural areas, forests, wetlands, and water. A sample of the HRSCD dataset is shown in Figure \ref{fig:hrscd}. Change labels can be easily acquired by comparing the land cover maps of bitemporal images. Since the HRSCD dataset provides complete bitemporal semantic segmentation labels, we can constrain the feature learning of the change detection task with multitask learning \cite{daudt2019multitask}. Semantic segmentation of bitemporal images can be introduced as an auxiliary task. In this case, the SSL-based strategy would be unnecessary. Therefore, we will not describe the experiment on this dataset in more detail. Those interested can refer to \cite{daudt2019multitask}.

\subsection{Comparison of the SSL-based strategy and contrastive loss strategy}
\label{S:5.2}

Since both the SSL-based strategy and contrastive loss strategy are used for constraining feature learning in an unsupervised way, we discuss the differences between these two methods in this subsection. The main difference lies in the fact that the SSL-based strategy uses a stricter constraint. For the contrastive loss strategy, the network needs to learn discriminative features. Unchanged features are pulled together, and changed features are pushed apart. However, ground objects in the bitemporal images are not distinguished. Compared with the contrastive loss strategy, the SSL-based strategy introduces an additional constraint that assigns bitemporal features to different semantics classes. With the supervision of additional semantic segmentation branches, changed features and unchanged features can be easily acquired by comparing bitemporal features.

We test the contrastive loss strategy and SSL-based strategy on the LEVIR-CD dataset and the SECON dataset. The comparison results are shown in Table \ref{Table:levir_second_acc}. From the table, we can learn that both the contrastive loss strategy and SSL-based strategy can boost model performance. Compared with the contrastive loss strategy, the SSL-based strategy performs better. On the LEVIR-CD dataset, SSL improves 0.29\% in terms of the IoU and 0.16\% in terms of the F1 score. On the SECOND dataset, SSL improves 0.2\% in terms of the mIoU and 0.24\% in terms of the SeK.

\begin{table}[ht]
\centering
\begin{tabular}{l c c c c c c}
\hline
\textbf{} & \multicolumn{4}{c|}{\textbf{LEVIR-CD}} & \multicolumn{2}{c}{\textbf{SECOND}} \\
\hline
DED & 92.88 & 90.67 & 84.78 & 91.76 & 70.34 & 19.17 \\
DED + ContraLoss & 92.84 & 90.98 & 85.01 & 91.90 & 70.77 & 19.86 \\
DED + SSL & \textbf{92.95} & \textbf{91.20} & \textbf{85.30} & \textbf{92.06} & \textbf{90.97} & \textbf{20.10} \\
\hline
\end{tabular}
\caption{Experimental result on the LEVIR-CD dataset and the SECOND dataset. DED in the table represents the dual encode-decode backbone proposed in Section \ref{S:3.2}. ContraLoss represents the contrastive loss strategy. SSL represents the SSL-based strategy.}
\label{Table:levir_second_acc}
\end{table}

\section{Conclusions}
\label{S:6}

In this paper, we propose a new change detection architecture for VHR remote sensing images. We design the algorithm for the purpose of extracting correct bitemporal features and fusing them in an effective way. To this end, we propose a DED backbone and an NL-FPN module to constrain the bitemporal feature extraction, DFM to fuse bitemporal features effectively, and an SSL-based strategy to constrain overall feature learning. Based on the above contributions, we propose FCCDN and validate it on two building change detection tasks. The experimental results show that FCCDN can achieve state-of-the-art performance with relatively high efficiency. Moreover, FCCDN can obtain bitemporal semantic segmentation results in an unsupervised way on the experimental datasets, which is vital for better Earth observations.

Although very promising, FCCDN still has several limitations. Firstly, as a supervised learning algorithm, FCCDN needs many labeled data to train a robust model. As building a change detection dataset is time-consuming, it is significant to seek solutions to change detection tasks with insufficient samples. Secondly, the SSL-based strategy may not work when applied to intra-class change detection tasks. If a ground object changes its appearance without changing its semantic class, the SSL-based strategy may not offer any additional supervision. In addition, since the SSL-based strategy relies on the supervision of the segmentation branches, it is essential to know the semantic categories of objects in bitemporal images. If the semantic categories are unknown, it is challenging to build the semantic segmentation brunches and formulate the auxiliary loss function. Consequently, future work will focus on fitting the proposed architecture to more datasets.

\section*{Acknowledgments}

The authors thank the editors and anonymous reviewers for their valuable comments, which greatly improved the quality of the paper.

This research was supported by the Strategic Priority Research Program of the Chinese Academy of Sciences under Grant No. XDA19080302.






\bibliographystyle{elsarticle-num-names}
\bibliography{FCCDN.bib}

\begin{thebibliography}{65}
\expandafter\ifx\csname natexlab\endcsname\relax\def\natexlab#1{#1}\fi
\providecommand{\url}[1]{\texttt{#1}}
\providecommand{\href}[2]{#2}
\providecommand{\path}[1]{#1}
\providecommand{\DOIprefix}{doi:}
\providecommand{\ArXivprefix}{arXiv:}
\providecommand{\URLprefix}{URL: }
\providecommand{\Pubmedprefix}{pmid:}
\providecommand{\doi}[1]{\href{http://dx.doi.org/#1}{\path{#1}}}
\providecommand{\Pubmed}[1]{\href{pmid:#1}{\path{#1}}}
\providecommand{\bibinfo}[2]{#2}
\ifx\xfnm\relax \def\xfnm[#1]{\unskip,\space#1}\fi
\bibitem[{MAHMOUDZADEH(2007)}]{mahmoudzadeh2007digital}
\bibinfo{author}{H.~MAHMOUDZADEH},
\newblock \bibinfo{title}{Digital change detection using remotely sensed data
  for monitoring green space destruction in tabriz}  (\bibinfo{year}{2007}).
\bibitem[{Shi et~al.(2020)Shi, Zhang, Zhang, Chen, and Zhan}]{shi2020change}
\bibinfo{author}{W.~Shi}, \bibinfo{author}{M.~Zhang},
  \bibinfo{author}{R.~Zhang}, \bibinfo{author}{S.~Chen},
  \bibinfo{author}{Z.~Zhan},
\newblock \bibinfo{title}{Change detection based on artificial intelligence:
  State-of-the-art and challenges},
\newblock \bibinfo{journal}{Remote Sensing} \bibinfo{volume}{12}
  (\bibinfo{year}{2020}) \bibinfo{pages}{1688}.
\bibitem[{Quarmby and Cushnie(1989)}]{quarmby1989monitoring}
\bibinfo{author}{N.~Quarmby}, \bibinfo{author}{J.~Cushnie},
\newblock \bibinfo{title}{Monitoring urban land cover changes at the urban
  fringe from spot hrv imagery in south-east england},
\newblock \bibinfo{journal}{International Journal of Remote Sensing}
  \bibinfo{volume}{10} (\bibinfo{year}{1989}) \bibinfo{pages}{953--963}.
\bibitem[{Howarth and Wickware(1981)}]{howarth1981procedures}
\bibinfo{author}{P.~J. Howarth}, \bibinfo{author}{G.~M. Wickware},
\newblock \bibinfo{title}{Procedures for change detection using landsat digital
  data},
\newblock \bibinfo{journal}{International Journal of Remote Sensing}
  \bibinfo{volume}{2} (\bibinfo{year}{1981}) \bibinfo{pages}{277--291}.
\bibitem[{Liu et~al.(2015)Liu, Bruzzone, Bovolo, Zanetti, and
  Du}]{liu2015sequential}
\bibinfo{author}{S.~Liu}, \bibinfo{author}{L.~Bruzzone},
  \bibinfo{author}{F.~Bovolo}, \bibinfo{author}{M.~Zanetti},
  \bibinfo{author}{P.~Du},
\newblock \bibinfo{title}{Sequential spectral change vector analysis for
  iteratively discovering and detecting multiple changes in hyperspectral
  images},
\newblock \bibinfo{journal}{IEEE Transactions on Geoscience and Remote Sensing}
  \bibinfo{volume}{53} (\bibinfo{year}{2015}) \bibinfo{pages}{4363--4378}.
\bibitem[{Richards(1984)}]{richards1984thematic}
\bibinfo{author}{J.~Richards},
\newblock \bibinfo{title}{Thematic mapping from multitemporal image data using
  the principal components transformation},
\newblock \bibinfo{journal}{Remote Sensing of Environment} \bibinfo{volume}{16}
  (\bibinfo{year}{1984}) \bibinfo{pages}{35--46}.
\bibitem[{Jin and Sader(2005)}]{jin2005comparison}
\bibinfo{author}{S.~Jin}, \bibinfo{author}{S.~A. Sader},
\newblock \bibinfo{title}{Comparison of time series tasseled cap wetness and
  the normalized difference moisture index in detecting forest disturbances},
\newblock \bibinfo{journal}{Remote sensing of Environment} \bibinfo{volume}{94}
  (\bibinfo{year}{2005}) \bibinfo{pages}{364--372}.
\bibitem[{Ghosh et~al.(2011)Ghosh, Mishra, and Ghosh}]{ghosh2011fuzzy}
\bibinfo{author}{A.~Ghosh}, \bibinfo{author}{N.~S. Mishra},
  \bibinfo{author}{S.~Ghosh},
\newblock \bibinfo{title}{Fuzzy clustering algorithms for unsupervised change
  detection in remote sensing images},
\newblock \bibinfo{journal}{Information Sciences} \bibinfo{volume}{181}
  (\bibinfo{year}{2011}) \bibinfo{pages}{699--715}.
\bibitem[{Im and Jensen(2005)}]{im2005change}
\bibinfo{author}{J.~Im}, \bibinfo{author}{J.~R. Jensen},
\newblock \bibinfo{title}{A change detection model based on neighborhood
  correlation image analysis and decision tree classification},
\newblock \bibinfo{journal}{Remote Sensing of Environment} \bibinfo{volume}{99}
  (\bibinfo{year}{2005}) \bibinfo{pages}{326--340}.
\bibitem[{Liu et~al.(2019)Liu, Chen, Xu, Sun, Yan, Diao, and
  Han}]{liu2019convolutional}
\bibinfo{author}{J.~Liu}, \bibinfo{author}{K.~Chen}, \bibinfo{author}{G.~Xu},
  \bibinfo{author}{X.~Sun}, \bibinfo{author}{M.~Yan},
  \bibinfo{author}{W.~Diao}, \bibinfo{author}{H.~Han},
\newblock \bibinfo{title}{Convolutional neural network-based transfer learning
  for optical aerial images change detection},
\newblock \bibinfo{journal}{IEEE Geoscience and Remote Sensing Letters}
  \bibinfo{volume}{17} (\bibinfo{year}{2019}) \bibinfo{pages}{127--131}.
\bibitem[{Cui et~al.(2019)Cui, Zhang, Yan, Wei, Wu
  et~al.}]{cui2019unsupervised}
\bibinfo{author}{B.~Cui}, \bibinfo{author}{Y.~Zhang}, \bibinfo{author}{L.~Yan},
  \bibinfo{author}{J.~Wei}, \bibinfo{author}{H.~Wu}, et~al.,
\newblock \bibinfo{title}{An unsupervised sar change detection method based on
  stochastic subspace ensemble learning},
\newblock \bibinfo{journal}{Remote Sensing} \bibinfo{volume}{11}
  (\bibinfo{year}{2019}) \bibinfo{pages}{1314}.
\bibitem[{LeCun et~al.(2015)LeCun, Bengio, and Hinton}]{lecun2015deep}
\bibinfo{author}{Y.~LeCun}, \bibinfo{author}{Y.~Bengio},
  \bibinfo{author}{G.~Hinton},
\newblock \bibinfo{title}{Deep learning},
\newblock \bibinfo{journal}{nature} \bibinfo{volume}{521}
  (\bibinfo{year}{2015}) \bibinfo{pages}{436--444}.
\bibitem[{Zhang et~al.(2019)Zhang, Chen, Peng, Benediktsson, Liu, Zou, Li, and
  Plaza}]{zhang2019remotely}
\bibinfo{author}{B.~Zhang}, \bibinfo{author}{Z.~Chen},
  \bibinfo{author}{D.~Peng}, \bibinfo{author}{J.~A. Benediktsson},
  \bibinfo{author}{B.~Liu}, \bibinfo{author}{L.~Zou}, \bibinfo{author}{J.~Li},
  \bibinfo{author}{A.~Plaza},
\newblock \bibinfo{title}{Remotely sensed big data: Evolution in model
  development for information extraction [point of view]},
\newblock \bibinfo{journal}{Proceedings of the IEEE} \bibinfo{volume}{107}
  (\bibinfo{year}{2019}) \bibinfo{pages}{2294--2301}.
\bibitem[{Daudt et~al.(2018)Daudt, Le~Saux, and Boulch}]{daudt2018fully}
\bibinfo{author}{R.~C. Daudt}, \bibinfo{author}{B.~Le~Saux},
  \bibinfo{author}{A.~Boulch},
\newblock \bibinfo{title}{Fully convolutional siamese networks for change
  detection},
\newblock in: \bibinfo{booktitle}{2018 25th IEEE International Conference on
  Image Processing (ICIP)}, \bibinfo{organization}{IEEE}, \bibinfo{year}{2018},
  pp. \bibinfo{pages}{4063--4067}.
\bibitem[{Peng et~al.(2019)Peng, Zhang, and Guan}]{peng2019end}
\bibinfo{author}{D.~Peng}, \bibinfo{author}{Y.~Zhang},
  \bibinfo{author}{H.~Guan},
\newblock \bibinfo{title}{End-to-end change detection for high resolution
  satellite images using improved unet++},
\newblock \bibinfo{journal}{Remote Sensing} \bibinfo{volume}{11}
  (\bibinfo{year}{2019}) \bibinfo{pages}{1382}.
\bibitem[{Zhang et~al.(2020)Zhang, Yue, Tapete, Jiang, Shangguan, Huang, and
  Liu}]{zhang2020deeply}
\bibinfo{author}{C.~Zhang}, \bibinfo{author}{P.~Yue},
  \bibinfo{author}{D.~Tapete}, \bibinfo{author}{L.~Jiang},
  \bibinfo{author}{B.~Shangguan}, \bibinfo{author}{L.~Huang},
  \bibinfo{author}{G.~Liu},
\newblock \bibinfo{title}{A deeply supervised image fusion network for change
  detection in high resolution bi-temporal remote sensing images},
\newblock \bibinfo{journal}{ISPRS Journal of Photogrammetry and Remote Sensing}
  \bibinfo{volume}{166} (\bibinfo{year}{2020}) \bibinfo{pages}{183--200}.
\bibitem[{Zhang et~al.(2021)Zhang, Hu, Zhang, Shu, and Zhou}]{zhang2021object}
\bibinfo{author}{L.~Zhang}, \bibinfo{author}{X.~Hu},
  \bibinfo{author}{M.~Zhang}, \bibinfo{author}{Z.~Shu},
  \bibinfo{author}{H.~Zhou},
\newblock \bibinfo{title}{Object-level change detection with a dual correlation
  attention-guided detector},
\newblock \bibinfo{journal}{ISPRS Journal of Photogrammetry and Remote Sensing}
  \bibinfo{volume}{177} (\bibinfo{year}{2021}) \bibinfo{pages}{147--160}.
\bibitem[{Zheng et~al.(2021)Zheng, Wan, Zhang, Xiang, Peng, and
  Zhang}]{zheng2021clnet}
\bibinfo{author}{Z.~Zheng}, \bibinfo{author}{Y.~Wan},
  \bibinfo{author}{Y.~Zhang}, \bibinfo{author}{S.~Xiang},
  \bibinfo{author}{D.~Peng}, \bibinfo{author}{B.~Zhang},
\newblock \bibinfo{title}{Clnet: Cross-layer convolutional neural network for
  change detection in optical remote sensing imagery},
\newblock \bibinfo{journal}{ISPRS Journal of Photogrammetry and Remote Sensing}
  \bibinfo{volume}{175} (\bibinfo{year}{2021}) \bibinfo{pages}{247--267}.
\bibitem[{Hou et~al.(2021)Hou, Bai, Li, Shang, and Shen}]{hou2021high}
\bibinfo{author}{X.~Hou}, \bibinfo{author}{Y.~Bai}, \bibinfo{author}{Y.~Li},
  \bibinfo{author}{C.~Shang}, \bibinfo{author}{Q.~Shen},
\newblock \bibinfo{title}{High-resolution triplet network with dynamic
  multiscale feature for change detection on satellite images},
\newblock \bibinfo{journal}{ISPRS Journal of Photogrammetry and Remote Sensing}
  \bibinfo{volume}{177} (\bibinfo{year}{2021}) \bibinfo{pages}{103--115}.
\bibitem[{Zhang et~al.(2021)Zhang, Fu, Li, and Zhang}]{zhang2021hdfnet}
\bibinfo{author}{Y.~Zhang}, \bibinfo{author}{L.~Fu}, \bibinfo{author}{Y.~Li},
  \bibinfo{author}{Y.~Zhang},
\newblock \bibinfo{title}{Hdfnet: Hierarchical dynamic fusion network for
  change detection in optical aerial images},
\newblock \bibinfo{journal}{Remote Sensing} \bibinfo{volume}{13}
  (\bibinfo{year}{2021}) \bibinfo{pages}{1440}.
\bibitem[{Zagoruyko and Komodakis(2015)}]{zagoruyko2015learning}
\bibinfo{author}{S.~Zagoruyko}, \bibinfo{author}{N.~Komodakis},
\newblock \bibinfo{title}{Learning to compare image patches via convolutional
  neural networks},
\newblock in: \bibinfo{booktitle}{Proceedings of the IEEE conference on
  computer vision and pattern recognition}, \bibinfo{year}{2015}, pp.
  \bibinfo{pages}{4353--4361}.
\bibitem[{Gao et~al.(2019)Gao, Gao, Dong, and Wang}]{gao2019change}
\bibinfo{author}{Y.~Gao}, \bibinfo{author}{F.~Gao}, \bibinfo{author}{J.~Dong},
  \bibinfo{author}{S.~Wang},
\newblock \bibinfo{title}{Change detection from synthetic aperture radar images
  based on channel weighting-based deep cascade network},
\newblock \bibinfo{journal}{IEEE Journal of Selected Topics in Applied Earth
  Observations and Remote Sensing} \bibinfo{volume}{12} (\bibinfo{year}{2019})
  \bibinfo{pages}{4517--4529}.
\bibitem[{Long et~al.(2015)Long, Shelhamer, and Darrell}]{long2015fully}
\bibinfo{author}{J.~Long}, \bibinfo{author}{E.~Shelhamer},
  \bibinfo{author}{T.~Darrell},
\newblock \bibinfo{title}{Fully convolutional networks for semantic
  segmentation},
\newblock in: \bibinfo{booktitle}{Proceedings of the IEEE conference on
  computer vision and pattern recognition}, \bibinfo{year}{2015}, pp.
  \bibinfo{pages}{3431--3440}.
\bibitem[{Alcantarilla et~al.(2018)Alcantarilla, Stent, Ros, Arroyo, and
  Gherardi}]{alcantarilla2018street}
\bibinfo{author}{P.~F. Alcantarilla}, \bibinfo{author}{S.~Stent},
  \bibinfo{author}{G.~Ros}, \bibinfo{author}{R.~Arroyo},
  \bibinfo{author}{R.~Gherardi},
\newblock \bibinfo{title}{Street-view change detection with deconvolutional
  networks},
\newblock \bibinfo{journal}{Autonomous Robots} \bibinfo{volume}{42}
  (\bibinfo{year}{2018}) \bibinfo{pages}{1301--1322}.
\bibitem[{Fang et~al.(2021)Fang, Li, Shao, and Li}]{fang2021snunet}
\bibinfo{author}{S.~Fang}, \bibinfo{author}{K.~Li}, \bibinfo{author}{J.~Shao},
  \bibinfo{author}{Z.~Li},
\newblock \bibinfo{title}{Snunet-cd: A densely connected siamese network for
  change detection of vhr images},
\newblock \bibinfo{journal}{IEEE Geoscience and Remote Sensing Letters}
  (\bibinfo{year}{2021}).
\bibitem[{Diakogiannis et~al.(2020)Diakogiannis, Waldner, and
  Caccetta}]{diakogiannis2020looking}
\bibinfo{author}{F.~I. Diakogiannis}, \bibinfo{author}{F.~Waldner},
  \bibinfo{author}{P.~Caccetta},
\newblock \bibinfo{title}{Looking for change? roll the dice and demand
  attention},
\newblock \bibinfo{journal}{arXiv preprint arXiv:2009.02062}
  (\bibinfo{year}{2020}).
\bibitem[{Hou et~al.(2019)Hou, Liu, Wang, and Wang}]{hou2019w}
\bibinfo{author}{B.~Hou}, \bibinfo{author}{Q.~Liu}, \bibinfo{author}{H.~Wang},
  \bibinfo{author}{Y.~Wang},
\newblock \bibinfo{title}{From w-net to cdgan: Bitemporal change detection via
  deep learning techniques},
\newblock \bibinfo{journal}{IEEE Transactions on Geoscience and Remote Sensing}
  \bibinfo{volume}{58} (\bibinfo{year}{2019}) \bibinfo{pages}{1790--1802}.
\bibitem[{Chen and Shi(2020)}]{chen2020spatial}
\bibinfo{author}{H.~Chen}, \bibinfo{author}{Z.~Shi},
\newblock \bibinfo{title}{A spatial-temporal attention-based method and a new
  dataset for remote sensing image change detection},
\newblock \bibinfo{journal}{Remote Sensing} \bibinfo{volume}{12}
  (\bibinfo{year}{2020}) \bibinfo{pages}{1662}.
\bibitem[{Ji et~al.(2018)Ji, Wei, and Lu}]{ji2018fully}
\bibinfo{author}{S.~Ji}, \bibinfo{author}{S.~Wei}, \bibinfo{author}{M.~Lu},
\newblock \bibinfo{title}{Fully convolutional networks for multisource building
  extraction from an open aerial and satellite imagery data set},
\newblock \bibinfo{journal}{IEEE Transactions on Geoscience and Remote Sensing}
  \bibinfo{volume}{57} (\bibinfo{year}{2018}) \bibinfo{pages}{574--586}.
\bibitem[{Peng et~al.(2020)Peng, Zhong, Li, and Li}]{peng2020optical}
\bibinfo{author}{X.~Peng}, \bibinfo{author}{R.~Zhong}, \bibinfo{author}{Z.~Li},
  \bibinfo{author}{Q.~Li},
\newblock \bibinfo{title}{Optical remote sensing image change detection based
  on attention mechanism and image difference},
\newblock \bibinfo{journal}{IEEE Transactions on Geoscience and Remote Sensing}
   (\bibinfo{year}{2020}).
\bibitem[{Liu et~al.(2020)Liu, Jiang, Zhang, and Zhang}]{liu2020deep}
\bibinfo{author}{R.~Liu}, \bibinfo{author}{D.~Jiang},
  \bibinfo{author}{L.~Zhang}, \bibinfo{author}{Z.~Zhang},
\newblock \bibinfo{title}{Deep depthwise separable convolutional network for
  change detection in optical aerial images},
\newblock \bibinfo{journal}{IEEE Journal of Selected Topics in Applied Earth
  Observations and Remote Sensing} \bibinfo{volume}{13} (\bibinfo{year}{2020})
  \bibinfo{pages}{1109--1118}.
\bibitem[{Zhou et~al.(2018)Zhou, Siddiquee, Tajbakhsh, and
  Liang}]{zhou2018unet++}
\bibinfo{author}{Z.~Zhou}, \bibinfo{author}{M.~M.~R. Siddiquee},
  \bibinfo{author}{N.~Tajbakhsh}, \bibinfo{author}{J.~Liang},
\newblock \bibinfo{title}{Unet++: A nested u-net architecture for medical image
  segmentation},
\newblock in: \bibinfo{booktitle}{Deep learning in medical image analysis and
  multimodal learning for clinical decision support},
  \bibinfo{publisher}{Springer}, \bibinfo{year}{2018}, pp.
  \bibinfo{pages}{3--11}.
\bibitem[{Ronneberger et~al.(2015)Ronneberger, Fischer, and
  Brox}]{ronneberger2015u}
\bibinfo{author}{O.~Ronneberger}, \bibinfo{author}{P.~Fischer},
  \bibinfo{author}{T.~Brox},
\newblock \bibinfo{title}{U-net: Convolutional networks for biomedical image
  segmentation},
\newblock in: \bibinfo{booktitle}{International Conference on Medical image
  computing and computer-assisted intervention},
  \bibinfo{organization}{Springer}, \bibinfo{year}{2015}, pp.
  \bibinfo{pages}{234--241}.
\bibitem[{Chen et~al.(2020)Chen, Yuan, Peng, Chen, Haozhe, Zhu, Liu, and
  Li}]{chen2020dasnet}
\bibinfo{author}{J.~Chen}, \bibinfo{author}{Z.~Yuan},
  \bibinfo{author}{J.~Peng}, \bibinfo{author}{L.~Chen},
  \bibinfo{author}{H.~Haozhe}, \bibinfo{author}{J.~Zhu},
  \bibinfo{author}{Y.~Liu}, \bibinfo{author}{H.~Li},
\newblock \bibinfo{title}{Dasnet: Dual attentive fully convolutional siamese
  networks for change detection of high resolution satellite images},
\newblock \bibinfo{journal}{IEEE Journal of Selected Topics in Applied Earth
  Observations and Remote Sensing}  (\bibinfo{year}{2020}).
\bibitem[{Jing and Tian(2020)}]{jing2020self}
\bibinfo{author}{L.~Jing}, \bibinfo{author}{Y.~Tian},
\newblock \bibinfo{title}{Self-supervised visual feature learning with deep
  neural networks: A survey},
\newblock \bibinfo{journal}{IEEE Transactions on Pattern Analysis and Machine
  Intelligence}  (\bibinfo{year}{2020}).
\bibitem[{Zhang et~al.(2016)Zhang, Isola, and Efros}]{zhang2016colorful}
\bibinfo{author}{R.~Zhang}, \bibinfo{author}{P.~Isola}, \bibinfo{author}{A.~A.
  Efros},
\newblock \bibinfo{title}{Colorful image colorization},
\newblock in: \bibinfo{booktitle}{European conference on computer vision},
  \bibinfo{organization}{Springer}, \bibinfo{year}{2016}, pp.
  \bibinfo{pages}{649--666}.
\bibitem[{Pathak et~al.(2016)Pathak, Krahenbuhl, Donahue, Darrell, and
  Efros}]{pathak2016context}
\bibinfo{author}{D.~Pathak}, \bibinfo{author}{P.~Krahenbuhl},
  \bibinfo{author}{J.~Donahue}, \bibinfo{author}{T.~Darrell},
  \bibinfo{author}{A.~A. Efros},
\newblock \bibinfo{title}{Context encoders: Feature learning by inpainting},
\newblock in: \bibinfo{booktitle}{Proceedings of the IEEE conference on
  computer vision and pattern recognition}, \bibinfo{year}{2016}, pp.
  \bibinfo{pages}{2536--2544}.
\bibitem[{Feng et~al.(2019)Feng, Xu, and Tao}]{feng2019self}
\bibinfo{author}{Z.~Feng}, \bibinfo{author}{C.~Xu}, \bibinfo{author}{D.~Tao},
\newblock \bibinfo{title}{Self-supervised representation learning by rotation
  feature decoupling},
\newblock in: \bibinfo{booktitle}{Proceedings of the IEEE/CVF Conference on
  Computer Vision and Pattern Recognition}, \bibinfo{year}{2019}, pp.
  \bibinfo{pages}{10364--10374}.
\bibitem[{Noroozi and Favaro(2016)}]{noroozi2016unsupervised}
\bibinfo{author}{M.~Noroozi}, \bibinfo{author}{P.~Favaro},
\newblock \bibinfo{title}{Unsupervised learning of visual representations by
  solving jigsaw puzzles},
\newblock in: \bibinfo{booktitle}{European conference on computer vision},
  \bibinfo{organization}{Springer}, \bibinfo{year}{2016}, pp.
  \bibinfo{pages}{69--84}.
\bibitem[{Tao et~al.(2020)Tao, Qi, Lu, Wang, and Li}]{tao2020remote}
\bibinfo{author}{C.~Tao}, \bibinfo{author}{J.~Qi}, \bibinfo{author}{W.~Lu},
  \bibinfo{author}{H.~Wang}, \bibinfo{author}{H.~Li},
\newblock \bibinfo{title}{Remote sensing image scene classification with
  self-supervised paradigm under limited labeled samples},
\newblock \bibinfo{journal}{IEEE Geoscience and Remote Sensing Letters}
  (\bibinfo{year}{2020}).
\bibitem[{Vincenzi et~al.(2020)Vincenzi, Porrello, Buzzega, Cipriano, Fronte,
  Cuccu, Ippoliti, Conte, and Calderara}]{vincenzi2020color}
\bibinfo{author}{S.~Vincenzi}, \bibinfo{author}{A.~Porrello},
  \bibinfo{author}{P.~Buzzega}, \bibinfo{author}{M.~Cipriano},
  \bibinfo{author}{P.~Fronte}, \bibinfo{author}{R.~Cuccu},
  \bibinfo{author}{C.~Ippoliti}, \bibinfo{author}{A.~Conte},
  \bibinfo{author}{S.~Calderara},
\newblock \bibinfo{title}{The color out of space: learning self-supervised
  representations for earth observation imagery},
\newblock \bibinfo{journal}{arXiv preprint arXiv:2006.12119}
  (\bibinfo{year}{2020}).
\bibitem[{Dong et~al.(2020)Dong, Ma, Wu, Zhang, and Jiao}]{dong2020self}
\bibinfo{author}{H.~Dong}, \bibinfo{author}{W.~Ma}, \bibinfo{author}{Y.~Wu},
  \bibinfo{author}{J.~Zhang}, \bibinfo{author}{L.~Jiao},
\newblock \bibinfo{title}{Self-supervised representation learning for remote
  sensing image change detection based on temporal prediction},
\newblock \bibinfo{journal}{Remote Sensing} \bibinfo{volume}{12}
  (\bibinfo{year}{2020}) \bibinfo{pages}{1868}.
\bibitem[{Chen and Bruzzone(2021)}]{chen2021self}
\bibinfo{author}{Y.~Chen}, \bibinfo{author}{L.~Bruzzone},
\newblock \bibinfo{title}{Self-supervised change detection in multi-view remote
  sensing images},
\newblock \bibinfo{journal}{arXiv preprint arXiv:2103.05969}
  (\bibinfo{year}{2021}).
\bibitem[{Hu et~al.(2018)Hu, Shen, and Sun}]{hu2018squeeze}
\bibinfo{author}{J.~Hu}, \bibinfo{author}{L.~Shen}, \bibinfo{author}{G.~Sun},
\newblock \bibinfo{title}{Squeeze-and-excitation networks},
\newblock in: \bibinfo{booktitle}{Proceedings of the IEEE conference on
  computer vision and pattern recognition}, \bibinfo{year}{2018}, pp.
  \bibinfo{pages}{7132--7141}.
\bibitem[{Ioffe and Szegedy(2015)}]{ioffe2015batch}
\bibinfo{author}{S.~Ioffe}, \bibinfo{author}{C.~Szegedy},
\newblock \bibinfo{title}{Batch normalization: Accelerating deep network
  training by reducing internal covariate shift},
\newblock in: \bibinfo{booktitle}{International conference on machine
  learning}, \bibinfo{organization}{PMLR}, \bibinfo{year}{2015}, pp.
  \bibinfo{pages}{448--456}.
\bibitem[{Selvaraju et~al.(2017)Selvaraju, Cogswell, Das, Vedantam, Parikh, and
  Batra}]{selvaraju2017grad}
\bibinfo{author}{R.~R. Selvaraju}, \bibinfo{author}{M.~Cogswell},
  \bibinfo{author}{A.~Das}, \bibinfo{author}{R.~Vedantam},
  \bibinfo{author}{D.~Parikh}, \bibinfo{author}{D.~Batra},
\newblock \bibinfo{title}{Grad-cam: Visual explanations from deep networks via
  gradient-based localization},
\newblock in: \bibinfo{booktitle}{Proceedings of the IEEE international
  conference on computer vision}, \bibinfo{year}{2017}, pp.
  \bibinfo{pages}{618--626}.
\bibitem[{Zhao et~al.(2017)Zhao, Shi, Qi, Wang, and Jia}]{zhao2017pyramid}
\bibinfo{author}{H.~Zhao}, \bibinfo{author}{J.~Shi}, \bibinfo{author}{X.~Qi},
  \bibinfo{author}{X.~Wang}, \bibinfo{author}{J.~Jia},
\newblock \bibinfo{title}{Pyramid scene parsing network},
\newblock in: \bibinfo{booktitle}{Proceedings of the IEEE conference on
  computer vision and pattern recognition}, \bibinfo{year}{2017}, pp.
  \bibinfo{pages}{2881--2890}.
\bibitem[{Diakogiannis et~al.(2020)Diakogiannis, Waldner, Caccetta, and
  Wu}]{diakogiannis2020resunet}
\bibinfo{author}{F.~I. Diakogiannis}, \bibinfo{author}{F.~Waldner},
  \bibinfo{author}{P.~Caccetta}, \bibinfo{author}{C.~Wu},
\newblock \bibinfo{title}{Resunet-a: a deep learning framework for semantic
  segmentation of remotely sensed data},
\newblock \bibinfo{journal}{ISPRS Journal of Photogrammetry and Remote Sensing}
  \bibinfo{volume}{162} (\bibinfo{year}{2020}) \bibinfo{pages}{94--114}.
\bibitem[{Chen et~al.(2017)Chen, Papandreou, Schroff, and
  Adam}]{chen2017rethinking}
\bibinfo{author}{L.-C. Chen}, \bibinfo{author}{G.~Papandreou},
  \bibinfo{author}{F.~Schroff}, \bibinfo{author}{H.~Adam},
\newblock \bibinfo{title}{Rethinking atrous convolution for semantic image
  segmentation},
\newblock \bibinfo{journal}{arXiv preprint arXiv:1706.05587}
  (\bibinfo{year}{2017}).
\bibitem[{Li et~al.(2018)Li, Xiong, An, and Wang}]{li2018pyramid}
\bibinfo{author}{H.~Li}, \bibinfo{author}{P.~Xiong}, \bibinfo{author}{J.~An},
  \bibinfo{author}{L.~Wang},
\newblock \bibinfo{title}{Pyramid attention network for semantic segmentation},
\newblock \bibinfo{journal}{arXiv preprint arXiv:1805.10180}
  (\bibinfo{year}{2018}).
\bibitem[{Lin et~al.(2017)Lin, Doll{\'a}r, Girshick, He, Hariharan, and
  Belongie}]{lin2017feature}
\bibinfo{author}{T.-Y. Lin}, \bibinfo{author}{P.~Doll{\'a}r},
  \bibinfo{author}{R.~Girshick}, \bibinfo{author}{K.~He},
  \bibinfo{author}{B.~Hariharan}, \bibinfo{author}{S.~Belongie},
\newblock \bibinfo{title}{Feature pyramid networks for object detection},
\newblock in: \bibinfo{booktitle}{Proceedings of the IEEE conference on
  computer vision and pattern recognition}, \bibinfo{year}{2017}, pp.
  \bibinfo{pages}{2117--2125}.
\bibitem[{Mi and Chen(2020)}]{mi2020superpixel}
\bibinfo{author}{L.~Mi}, \bibinfo{author}{Z.~Chen},
\newblock \bibinfo{title}{Superpixel-enhanced deep neural forest for remote
  sensing image semantic segmentation},
\newblock \bibinfo{journal}{ISPRS Journal of Photogrammetry and Remote Sensing}
  \bibinfo{volume}{159} (\bibinfo{year}{2020}) \bibinfo{pages}{140--152}.
\bibitem[{Wang et~al.(2018)Wang, Girshick, Gupta, and He}]{wang2018non}
\bibinfo{author}{X.~Wang}, \bibinfo{author}{R.~Girshick},
  \bibinfo{author}{A.~Gupta}, \bibinfo{author}{K.~He},
\newblock \bibinfo{title}{Non-local neural networks},
\newblock in: \bibinfo{booktitle}{Proceedings of the IEEE conference on
  computer vision and pattern recognition}, \bibinfo{year}{2018}, pp.
  \bibinfo{pages}{7794--7803}.
\bibitem[{Yuan and Wang(2018)}]{yuan2018ocnet}
\bibinfo{author}{Y.~Yuan}, \bibinfo{author}{J.~Wang},
\newblock \bibinfo{title}{Ocnet: Object context network for scene parsing},
\newblock \bibinfo{journal}{arXiv preprint arXiv:1809.00916}
  (\bibinfo{year}{2018}).
\bibitem[{Xu et~al.(2017)Xu, Ranftl, and Koltun}]{xu2017accurate}
\bibinfo{author}{J.~Xu}, \bibinfo{author}{R.~Ranftl},
  \bibinfo{author}{V.~Koltun},
\newblock \bibinfo{title}{Accurate optical flow via direct cost volume
  processing},
\newblock in: \bibinfo{booktitle}{Proceedings of the IEEE Conference on
  Computer Vision and Pattern Recognition}, \bibinfo{year}{2017}, pp.
  \bibinfo{pages}{1289--1297}.
\bibitem[{Zhou et~al.(2018)Zhou, Zhang, and Wu}]{zhou2018d}
\bibinfo{author}{L.~Zhou}, \bibinfo{author}{C.~Zhang}, \bibinfo{author}{M.~Wu},
\newblock \bibinfo{title}{D-linknet: Linknet with pretrained encoder and
  dilated convolution for high resolution satellite imagery road extraction},
\newblock in: \bibinfo{booktitle}{Proceedings of the IEEE Conference on
  Computer Vision and Pattern Recognition Workshops}, \bibinfo{year}{2018}, pp.
  \bibinfo{pages}{182--186}.
\bibitem[{Buslaev et~al.(2020)Buslaev, Iglovikov, Khvedchenya, Parinov,
  Druzhinin, and Kalinin}]{buslaev2020albumentations}
\bibinfo{author}{A.~Buslaev}, \bibinfo{author}{V.~I. Iglovikov},
  \bibinfo{author}{E.~Khvedchenya}, \bibinfo{author}{A.~Parinov},
  \bibinfo{author}{M.~Druzhinin}, \bibinfo{author}{A.~A. Kalinin},
\newblock \bibinfo{title}{Albumentations: fast and flexible image
  augmentations},
\newblock \bibinfo{journal}{Information} \bibinfo{volume}{11}
  (\bibinfo{year}{2020}) \bibinfo{pages}{125}.
\bibitem[{Paszke et~al.(2017)Paszke, Gross, Chintala, Chanan, Yang, DeVito,
  Lin, Desmaison, Antiga, and Lerer}]{paszke2017automatic}
\bibinfo{author}{A.~Paszke}, \bibinfo{author}{S.~Gross},
  \bibinfo{author}{S.~Chintala}, \bibinfo{author}{G.~Chanan},
  \bibinfo{author}{E.~Yang}, \bibinfo{author}{Z.~DeVito},
  \bibinfo{author}{Z.~Lin}, \bibinfo{author}{A.~Desmaison},
  \bibinfo{author}{L.~Antiga}, \bibinfo{author}{A.~Lerer},
\newblock \bibinfo{title}{Automatic differentiation in pytorch}
  (\bibinfo{year}{2017}).
\bibitem[{Kingma and Ba(2014)}]{kingma2014adam}
\bibinfo{author}{D.~P. Kingma}, \bibinfo{author}{J.~Ba},
\newblock \bibinfo{title}{Adam: A method for stochastic optimization},
\newblock \bibinfo{journal}{arXiv preprint arXiv:1412.6980}
  (\bibinfo{year}{2014}).
\bibitem[{Hendrycks et~al.(2019)Hendrycks, Lee, and
  Mazeika}]{hendrycks2019using}
\bibinfo{author}{D.~Hendrycks}, \bibinfo{author}{K.~Lee},
  \bibinfo{author}{M.~Mazeika},
\newblock \bibinfo{title}{Using pre-training can improve model robustness and
  uncertainty},
\newblock in: \bibinfo{booktitle}{International Conference on Machine
  Learning}, \bibinfo{organization}{PMLR}, \bibinfo{year}{2019}, pp.
  \bibinfo{pages}{2712--2721}.
\bibitem[{Chen et~al.(2021)Chen, Qi, and Shi}]{chen2021efficient}
\bibinfo{author}{H.~Chen}, \bibinfo{author}{Z.~Qi}, \bibinfo{author}{Z.~Shi},
\newblock \bibinfo{title}{Efficient transformer based method for remote sensing
  image change detection},
\newblock \bibinfo{journal}{arXiv preprint arXiv:2103.00208}
  (\bibinfo{year}{2021}).
\bibitem[{Papadomanolaki et~al.(2019)Papadomanolaki, Verma, Vakalopoulou,
  Gupta, and Karantzalos}]{papadomanolaki2019detecting}
\bibinfo{author}{M.~Papadomanolaki}, \bibinfo{author}{S.~Verma},
  \bibinfo{author}{M.~Vakalopoulou}, \bibinfo{author}{S.~Gupta},
  \bibinfo{author}{K.~Karantzalos},
\newblock \bibinfo{title}{Detecting urban changes with recurrent neural
  networks from multitemporal sentinel-2 data},
\newblock in: \bibinfo{booktitle}{IGARSS 2019-2019 IEEE International
  Geoscience and Remote Sensing Symposium}, \bibinfo{organization}{IEEE},
  \bibinfo{year}{2019}, pp. \bibinfo{pages}{214--217}.
\bibitem[{Lebedev et~al.(2018)Lebedev, Vizilter, Vygolov, Knyaz, and
  Rubis}]{lebedev2018change}
\bibinfo{author}{M.~Lebedev}, \bibinfo{author}{Y.~V. Vizilter},
  \bibinfo{author}{O.~Vygolov}, \bibinfo{author}{V.~Knyaz},
  \bibinfo{author}{A.~Y. Rubis},
\newblock \bibinfo{title}{Change detection in remote sensing images using
  conditional adversarial networks.},
\newblock \bibinfo{journal}{International Archives of the Photogrammetry,
  Remote Sensing \& Spatial Information Sciences} \bibinfo{volume}{42}
  (\bibinfo{year}{2018}).
\bibitem[{Yang et~al.(2020)Yang, Xia, Liu, Du, Yang, and
  Pelillo}]{yang2020asymmetric}
\bibinfo{author}{K.~Yang}, \bibinfo{author}{G.-S. Xia},
  \bibinfo{author}{Z.~Liu}, \bibinfo{author}{B.~Du}, \bibinfo{author}{W.~Yang},
  \bibinfo{author}{M.~Pelillo},
\newblock \bibinfo{title}{Asymmetric siamese networks for semantic change
  detection},
\newblock \bibinfo{journal}{arXiv preprint arXiv:2010.05687}
  (\bibinfo{year}{2020}).
\bibitem[{Daudt et~al.(2019)Daudt, Le~Saux, Boulch, and
  Gousseau}]{daudt2019multitask}
\bibinfo{author}{R.~C. Daudt}, \bibinfo{author}{B.~Le~Saux},
  \bibinfo{author}{A.~Boulch}, \bibinfo{author}{Y.~Gousseau},
\newblock \bibinfo{title}{Multitask learning for large-scale semantic change
  detection},
\newblock \bibinfo{journal}{Computer Vision and Image Understanding}
  \bibinfo{volume}{187} (\bibinfo{year}{2019}) \bibinfo{pages}{102783}.

\end{thebibliography}







\end{document}